%% file: final.tex
\documentclass{article} 
\usepackage{arxiv}

\input{preamble/preamble}

\title{Trajectory Replay in Sequential Flow
Matching: \\Concentration \& Coverage}

\author{
Karan J. Andrade \qquad Felipe Tobar  \vspace{0.5em} \\
Department of Mathematics\\Imperial College London \\
\texttt{\{karan.andrade25,f.tobar\}@imperial.ac.uk}
}

\begin{document}

\maketitle

\begin{abstract}
Recent work on sequential flow matching shows that the infinite-capacity empirical velocity field is a kernel regressor over stored transitions, approaching nearest-neighbour trajectory replay as its weights concentrate. We characterise when this phenomenon occurs and identify which modelling choices control it. For Gaussian conditional paths, we characterise the onset of replay, its behaviour in high dimensions, and how it depends on the geometry of the stored transitions and the choice of conditional path. We further show that, under constant-in-flow-time covariance, the empirical field remains within the convex hull of the stored velocity targets, imposing a representation floor that cannot be overcome by reweighting. Experiments on chaotic systems validate the predicted concentration and coverage behaviour, while finite-capacity experiments show that neural networks can depart from the empirical minimiser, the exact mechanism of which is left as an open question. Together, these results distinguish replay concentration from target coverage, thus informing future sequential flow matching modeling choices.
\end{abstract}

\input{Chapters/Introduction}

\input{Chapters/Preliminaries}

\input{Chapters/Occurence}

\input{Chapters/Concentration}

\input{Chapters/Coverage}

\input{Chapters/Empirical}

\section{Related Work}

Sequential flow matching and stochastic interpolants have been developed for
forecasting, conditional trajectories and multi-marginal temporal modelling
\citep{zhang2024,chen2024si,tamir2024conditional,lim2025paths, lee2025,islam2025,rohbeck2025,kviman2026}, while coupling design has been used
to simplify or straighten learned flows \citep{pooladian2023,tong2024}.
Our analysis instead studies the unrestricted empirical regression solution and
when its posterior weights concentrate onto stored sequential targets. This
connects the problem to classical high-dimensional nonparametric regression and
distance concentration \citep{beyer1999,bickel2007local,kpotufe2011k}, as well
as to recent work on memorisation and finite-sample behaviour in diffusion models
\citep{carlini2023extracting,somepalli2023diffusion,biroli2024,
achilli2024,kadkhodaie2024}. \citet{biroli2024} in particular conduct similar concentration analysis to our \cref{prop:logM}. \cref{fig:onset}(c) demonstrates logarithmic scaling, while the onset constants vary with bank geometry and the ESS criterion. They do not strictly follow the constant in \citet{biroli2024}. Our result is thus a broader but less sharp characterisation that suits the use case of sequential data. A more comprehensive comparison, including finite-capacity generalisation, is given in \cref{app:relworks}.

\section{Discussion and Limitations}

Our results revealed two distinct behaviours in the empirical sequential flow-matching estimator. Specifically, we identified replay concentration, which concerns how strongly the posterior weights select individual stored windows, and target coverage, which concerns whether the stored conditional targets can represent the required velocity at all. These mechanisms respond differently to modelling choices: window order and interpolation change the geometry of the stored centres and targets, while bandwidth and covariance geometry control concentration. We also showed that, under constant-in-flow-time covariance, changing these weights cannot move the empirical field outside the convex hull of the stored targets, confirming that reducing replay is not by itself a prescription for improving prediction. Our finite-capacity experiment, also revealed that wider networks inherit substantially more of the closed-form replay behaviour, while narrower networks can depart from this and achieve lower off-support error. 

Our analysis gives a starting point for two natural extensions. First, our concentration results characterise queries generated from a stored conditional component, which isolates the mechanism responsible for replay. Extending this analysis to off-support queries would connect the proposed weight-gap framework more directly to inference, where the query need not lie near any stored trajectory. Second, our coverage results show that, for constant-in-flow-time covariance,
reweighting alone cannot overcome deficiencies in the stored target set. The time-varying isotropic extension establishes a path-dependent representation floor; extending this characterisation to general matrix-valued covariance schedules remains open. This motivates the study of more general probability paths. Such extensions could clarify when modifying the path can improve not only concentration, but also the representational capacity of the empirical field.

\subsection*{AI use statement}

We used generative AI tools for literature review, formalisation of notations and definitions, refining the writing, coding tasks, and rendering the illustration in Figure \ref{fig:concentration_coverage_schematic}. We have not used generative AI tools for unsupervised generation of any mathematical statements or proofs in this paper. We have reviewed all AI-assisted code and figures. We take responsibility for the final content of this work, including text, claims or artefacts produced with the aid of generative AI.

\subsection*{Reproducibility statement}

The closed form experiments in \cref{sec:empirical-onset,sec:empirical-coverage,sec:empirical-anisotropy} evaluate the empirical flow-matching field directly; no network is trained. Randomness is controlled via fixed seeds defining the stored banks and evaluation queries, and comparisons within an experiment use the same banks and query sets. 

The finite capacity experiment in \cref{sec:empirical-neural} additionally fixes the training protocol and reports results across independent training seeds. Experiment-specific systems, sample counters, hyperparameters, aggregation procedures and implementation details are provided in \cref{app:empirical-details}. 

Code, configurations and scripts required to reproduce the experiments and figures are available at \url{https://github.com/kj-andrade/trajectory_replay_concentration_coverage}

\bibliography{iclr2027_conference}
\bibliographystyle{iclr2027_conference}

\newpage

\appendix

The appendix is structured in the following way:
\begin{enumerate}
    \item \cref{app:proofs} contains the proofs for every theorem and proposition stated in the main text, as well as a few addition remarks and corollaries. Each section has its proofs in \cref{app:proofs_sec3,app:proofs_sec4,app:proofs_sec5}.
    \item \cref{app:empirical-details} contains detailed description of the empirical tests conducted.
    \item \cref{app:relworks} contains a more detailed discussion of the project at large, including related works and future directions.
\end{enumerate}

\input{Chapters/Appendix}

\end{document}

%% file: preamble/preamble.tex
\usepackage{amsmath,amssymb,amsthm,mathtools}
\usepackage{booktabs}
\usepackage{graphicx}
\usepackage[hidelinks]{hyperref}\usepackage{url}
\usepackage{url}
\usepackage[nameinlink,capitalise]{cleveref}
\usepackage{xcolor}
\usepackage{placeins}
\usepackage{tikz}
\usetikzlibrary{arrows.meta,calc,positioning,fit,backgrounds}
\usepackage{subcaption}
\usepackage{natbib}

\definecolor{storedblue}{RGB}{55,100,135}
\definecolor{queryorange}{RGB}{220,95,35}
\definecolor{softblue}{RGB}{220,232,240}
\definecolor{softgray}{RGB}{245,245,245}

\newtheorem{theorem}{Theorem}
\newtheorem{proposition}{Proposition}
\newtheorem{corollary}{Corollary}
\newtheorem{lemma}{Lemma}
\newtheorem{remark}{Remark}
\newtheorem{definition}{Definition}

\DeclareMathOperator{\Cov}{Cov}
\DeclareMathOperator{\Var}{Var}
\DeclareMathOperator{\tr}{tr}
\DeclareMathOperator{\conv}{conv}

\newcommand{\R}{\mathbb{R}}
\newcommand{\E}{\mathbb{E}}
\newcommand{\Prob}{\mathbb{P}}
\newcommand{\Norm}{\mathcal{N}}
\newcommand{\ESS}{\mathrm{ESS}}
\newcommand{\reff}{r_{\mathrm{eff}}}

\newcommand{\ut}{u_t^\circ}

\newcommand{\mjt}{\mu_t^{(j)}}
\newcommand{\mti}{\mu_t^{(i)}}

\newcommand{\Pperp}{P^{\perp}}

%% file: Chapters/Introduction.tex
\section{Introduction}\label{chap:intro}

Flow matching and stochastic-interpolant formulations provide a regression-based
approach to learning flow and diffusion models \citep{lipman2023,tong2024,albergo2025stochastic,pooladian2023}. Their use for generating data with temporal structure can adopt diverse paradigms, with growing attention to modelling the evolution of sequences generated by dynamical systems \citep{zhang2024,chen2024si}. When conditioning on multiple consecutive states, multi-marginal approaches learn consistent paths across timestamps \citep{lee2025,islam2025,rohbeck2025,kviman2026}. In these settings, this work is concerned with the optimal velocity field obtained when the class of admissible models for such a field is unrestricted.

For canonical Gaussian conditional paths between states and a training set of state transitions, \citet{lim2026} show that the empirical minimiser, denoted $v^\star_t$, has a memory-based structure given by a weighted kernel estimator. This Nadaraya-Watson estimator is known to suffer from the curse of dimensionality, with convergence rates deteriorating with the state dimension \citep{stone1980}. An open question in the sequential setting is \emph{if/when} this estimator enters the nearest-neighbour regime (aka trajectory replay) and which modelling choices can mitigate such phenomenon. 

\begin{figure*}[t]
\centering

\resizebox{\textwidth}{!}{%
\begin{tikzpicture}[
    >=Latex,
    font=\small,
    every node/.style={align=center},
    stored/.style={storedblue, line width=1.1pt},
    target/.style={storedblue, -{Latex[length=2mm]}, line width=1.1pt},
    desired/.style={queryorange, -{Latex[length=2.4mm]}, line width=1.6pt},
]


\node[inner sep=0pt] (lorenz) at (2.0,2.15)
    {\includegraphics[width=3.7cm]{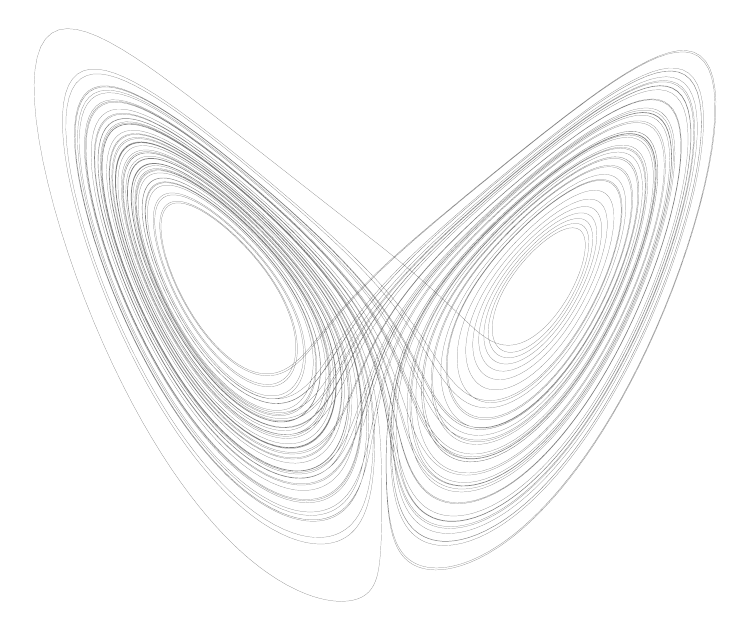}};

\node[font=\bfseries] at (2.0,4.15)
    {Lorenz--63 attractor};

\coordinate (boxSW) at ($(lorenz.center)+(-1.48,0.50)$);
\coordinate (boxNE) at ($(lorenz.center)+(-0.70,1.15)$);

\draw[queryorange, line width=1.2pt]
    (boxSW) rectangle (boxNE);

\node[queryorange, anchor=south, font=\scriptsize]
    at ($(boxNE)+(0,0.1)$)
    {local neighbourhood};


\begin{scope}[shift={(4.15,0.20)}]

\draw[rounded corners=3pt, line width=0.8pt]
    (0,0.55) rectangle (5.0,3.95);

\node[font=\bfseries, anchor=west]
    at (0.35,3.60) {Concentration};

\draw[stored]
    (0.35,2.75)
    .. controls (1.00,3.00) and (1.65,3.05) ..
    (2.60,3.20);

\draw[stored]
    (0.35,2.25)
    .. controls (1.05,2.45) and (1.75,2.48) ..
    (2.70,2.60);

\draw[stored]
    (0.45,1.72)
    .. controls (1.15,1.82) and (1.85,1.78) ..
    (2.70,1.58);

\coordinate (m1) at (1.20,3.00);
\coordinate (m2) at (1.70,2.48);
\coordinate (m3) at (2.05,1.70);

\filldraw[fill=white, draw=storedblue, line width=1.2pt]
    (m1) circle (2.1pt);
\filldraw[fill=white, draw=storedblue, line width=1.2pt]
    (m2) circle (2.1pt);
\filldraw[fill=white, draw=storedblue, line width=1.2pt]
    (m3) circle (2.1pt);

\coordinate (q) at (0.78,2.30);
\fill[queryorange] (q) circle (2.8pt);
\node[queryorange, anchor=east, font=\scriptsize]
    at ($(q)+(-0.05,0.1)$) {$x$};

\draw[storedblue, line width=2.0pt, opacity=.85]
    (q) -- (m2);
\draw[storedblue, line width=1.2pt, opacity=.65]
    (q) -- (m1);
\draw[storedblue, line width=0.55pt, opacity=.45]
    (q) -- (m3);

\node[font=\scriptsize, storedblue] at (1.22,2.20) {$\alpha_1$};
\node[font=\scriptsize, storedblue] at (1.25,2.69) {$\alpha_2$};
\node[font=\scriptsize, storedblue] at (1.35,1.85) {$\alpha_3$};

\draw[target] (m1) -- ++(0.42,0.16);
\draw[target] (m2) -- ++(0.46,0.10);
\draw[target] (m3) -- ++(0.44,-0.06);

\draw[-{Latex[length=2.5mm]}, line width=1.4pt]
    (q) -- ++(0.92,0.24);

\node[
    font=\scriptsize,
    text width=1.85cm,
    anchor=west
] at (2.55,2.45)
{
    \[
    v_t^\star(x)
    =
    \sum_j \alpha_j u_t^{(j)}
    \]
    Line thickness indicates the relative size of $\alpha_j$.
};

\end{scope}


\begin{scope}[shift={(9.45,0.20)}]

\draw[rounded corners=3pt, line width=0.8pt]
    (0,0.55) rectangle (5.15,3.95);

\node[font=\bfseries, anchor=west]
    at (0.35,3.60) {Target coverage};

\coordinate (o) at (0.65,1.55);

\coordinate (u1) at ($(o)+(0.58,1.05)$);
\coordinate (u2) at ($(o)+(1.35,0.88)$);
\coordinate (u3) at ($(o)+(1.55,0.28)$);
\coordinate (u4) at ($(o)+(0.78,0.02)$);

\fill[storedblue!13]
    (u1) -- (u2) -- (u3) -- (u4) -- cycle;

\draw[storedblue, dashed, line width=0.8pt]
    (u1) -- (u2) -- (u3) -- (u4) -- cycle;

\draw[target] (o) -- (u1);
\draw[target] (o) -- (u2);
\draw[target] (o) -- (u3);
\draw[target] (o) -- (u4);

\node[font=\scriptsize, anchor=north east]
    at ($(o)+(-0.02,-0.03)$) {$0$};

\coordinate (f) at ($(o)+(1.75,1.65)$);
\draw[desired] (o) -- (f);

\node[queryorange, font=\scriptsize, anchor=west]
    at ($(f)+(0.03,0.02)$) {$f_t(x)$};

\coordinate (closest) at ($(o)+(1.32,0.88)$);

\draw[queryorange, dashed, line width=0.9pt]
    (closest) -- (f);

\node[queryorange, font=\scriptsize]
    at (2.55,2.55) {$D_t(x)$};

\node[
    font=\scriptsize,
    text width=2.05cm,
    anchor=west
] at (2.85,2.05)
{
    $\mathcal U_t
    =
    \operatorname{conv}
    \{u_t^{(j)}\}$

    \vspace{0.15cm}

    Changing $\alpha$ moves the estimate inside $\mathcal U_t$, but cannot
    remove the representation floor.
};

\end{scope}


\draw[-{Latex[length=2.7mm]}, line width=1.0pt]
    ($(boxNE)+(0.05,-0.05)$)
    .. controls (3.30,3.15) and (3.65,3.65) ..
    (4.15,3.85);

\end{tikzpicture}
}

\caption{
Schematic distinction between \emph{concentration} and \emph{target coverage} questions. Orange arrow's represent the present query and the blue arrows represent stored transitions.
}
\label{fig:concentration_coverage_schematic}

\end{figure*}

Building on the multi-marginal extension of \citet{lim2026}, we formulate and address two distinct questions. \textbf{Concentration:} How does the flow-model specification control the transition of $v^\star_t$ from kernel averaging to nearest-neighbour replay? \textbf{Target coverage:} Can convex combinations of the stored conditional targets represent the required velocity at a given query? This distinction, illustrated geometrically in \cref{fig:concentration_coverage_schematic}, separates the mechanism producing replay from the question of whether replay, smoothing or any intermediate weighting scheme can be accurate.

Our contributions are as follows: (\romannumeral 1) We define a concentration parameter determined by the training set and probability path, and show that replay emerges when such a parameter grows above a threshold of order $\log M$, where $M$ is the number of stored state transitions. \citet{biroli2024} analyse a similar Gaussian-mixture situation, and establish a logarithmic scaling with sharp bounds for Gaussian-mixture weights under their distributional assumptions. Our bounds are geometry and memory-bank dependent, thus they do not require independent Gaussian centres. (\romannumeral 2) We characterise the effect of the probability path on replay concentration. (\romannumeral 3) For constant-covariance paths, we show that $v^\star_t$ lies in the convex hull of the stored velocity targets, establishing which velocities are unattainable by reweighting. (\romannumeral 4) Lastly, we validate these theoretical claims experimentally.

Our theoretical results concern the infinite-capacity (unrestricted) minimiser $v^\star_t$ of the empirical flow-matching objective for a fixed trajectory dataset. Accordingly, our experiments evaluate this closed-form field directly, with an additional finite-capacity experiment assessing the extent to which trained neural networks inherit its behaviour. We use chaotic systems as a primary testbed because their trajectories repeatedly visit the same regions without exactly repeating, making the distinction between trajectory replay and gaps in the stored trajectory bank particularly apparent.

%% file: Chapters/Preliminaries.tex
\section{Preliminaries}\label{chap:preliminaries}

\paragraph{Sequential flow matching and notation.}

Flow matching specifies a conditional probability path \(p_t(\cdot\mid Z)\) and velocity target \(u_t^\circ(x\mid Z)\), to learn a velocity field by minimising the loss given by
\[
\mathcal{L}_v(v)=\E_{t,Z,X_t}\!\left[\|v_t(X_t)-\ut(X_t\mid Z)\|^2\right],
\qquad X_t\sim p_t(\cdot\mid Z).
\]
For an unrestricted square-integrable class of velocity-field functions, the minimiser is the usual squared-loss regression solution \citep{lipman2023}, given by the conditional expectation
\[
v_t^\star(x)=\E[\ut(X_t\mid Z)\mid X_t=x].
\]
In sequential flow matching, the conditioning variable \(Z\) is constructed from observed states \(x_0,\ldots,x_T\in\R^d\). For a more general case, we define the conditioning variable to be a window, of order \(k\geq1\), as
\(
W_\tau=(x_\tau,\ldots,x_{\tau+k}),
\)
and use the empirical distribution over the \(M\) stored windows as the multi-marginal coupling,
\(
\hat\pi=\frac1M\sum_{\tau=1}^M\delta_{W_\tau}.
\)
Notice that \(\tau\) denotes physical or observation time, while \(t\in[0,1]\) denotes the flow's abstract time.

For an isotropic Gaussian conditional path
\(p_t(\cdot\mid W)=\Norm(\mu_t(W),\sigma_t^2I_d)\), where \(\mu_t(W)\) is specified by the chosen interpolant, the conditional velocity target is
\[
\ut(x\mid W)
=
\dot\mu_t(W)+\frac{\sigma_t'}{\sigma_t}\bigl(x-\mu_t(W)\bigr).
\]
For constant variance, \(\sigma_t'=0\), the target reduces to \(\ut(x\mid W)= \dot\mu_t(W)\).

\paragraph{Empirical minimiser and the \emph{trajectory replay} phenomenon.}

We state the empirical minimiser presented by \citet{lim2026}(Appendix B, Proposition 4).

\begin{lemma}[Empirical velocity minimiser \citep{lim2026}]
\label{lem:mainminimizer}
Fix \(t\in(0,1)\) and let \(\hat\pi=\frac1M\sum_{j=1}^M\delta_{Z_j}\) be the uniform
empirical coupling over the stored conditioning variables \(Z_1,\dots,Z_M\).
Suppose each conditional path \(p_t(\cdot\mid Z_j)\) has a density that is strictly
positive on \(\R^d\), and
\(\int p_t(x\mid Z_j)\,\|\ut(x\mid Z_j)\|^2\,dx<\infty\) for each \(j\).
Then, over square-integrable measurable fields, \(\mathcal L_v\) has the unique a.e. minimiser
\[
v_t^\star(x)=\sum_{j=1}^M\alpha_j(t,x)\,\ut(x\mid Z_j),
\qquad
\alpha_j(t,x)=\frac{p_t(x\mid Z_j)}{\sum_{l=1}^M p_t(x\mid Z_l)},
\]
that is, \(v_t^\star(x)=\E[\ut(X_t\mid Z)\mid X_t=x]\) with \(Z\sim\hat\pi\) and
\(X_t\mid Z\sim p_t(\cdot\mid Z)\).
\end{lemma}

\begin{remark}
For a Gaussian path $p_t$, the minimiser in \cref{lem:mainminimizer} takes the form of an explicit Gaussian kernel smoother. Since the weights $\{\alpha_j\}_{j=1}^M$ can concentrate on a single value, replay is intrinsically linked to the choice of conditional path (which includes modeling choices of the conditioning variables themselves). 
\end{remark}

Explicitly, for a Gaussian path $p_t(x\mid Z_j) \propto \exp\!\left(-\|x-\mu_t^{(j)}\|^2/(2\sigma_t^2)\right)$, where
\(\mu_t^{(j)}=\mu_t(Z_j)\), these weights become
\begin{equation}
    \alpha_j(t,x)
=
\frac{
\exp\!\left(-\|x-\mu_t^{(j)}\|^2/(2\sigma_t^2)\right)
}{
\sum_l
\exp\!\left(-\|x-\mu_t^{(l)}\|^2/(2\sigma_t^2)\right)
}.    
\label{eq:alphas}
\end{equation}

We refer to $\sigma_t^2$ as the kernel bandwidth. The conditioning variable $Z_j$ may be a pairwise transition or a higher-order window. Critically, window order and interpolation can change the interpolated path centres and conditional targets, while the similarity weight form of the estimator remains invariant.

%% file: Chapters/Occurence.tex
\section{When Trajectory Replay occurs} \label{chap:replay_section}

The proofs for all theorems and propositions in \cref{chap:replay_section,chap:replayconc,chap:targetspacecover} are provided in \cref{app:proofs} of the Appendix. 

We now ask when the weights in eq.~\eqref{eq:alphas} become concentrated enough for trajectory replay. Consider a query point $x\in \R^d$ (i.e., a state at which the empirical velocity field is evaluated) drawn from the conditional path associated with the stored window index $i$ as $x= \mti+\sigma_t\varepsilon, i\sim\mathrm{Unif}\{1,\dots,M\}, \varepsilon\sim\Norm(0,I_d)$. By concentration, we mean that a single weight in eq.~\eqref{eq:alphas} carries nearly all of the mass, therefore, replay corresponds to $\alpha_i(t,x)\approx1$ for window $W_i$.

\begin{definition}
    For the shared isotropic Gaussian path, and a pair of window indices \(i\) and \(j\), define the dimensionless concentration
    ratio
\begin{equation}
\label{eq:kappa}
    \kappa_{ij}\coloneqq\frac{\|\mti-\mjt\|^2}{\sigma_t^2},    
\end{equation}
which compares the squared separation between two path centres with the conditional path variance, and will be the basic quantity controlling concentration in our work.
\end{definition}

We first provide a pairwise comparison between windows, and then consider the full memory bank. We proceed by defining and analysing the log-weight gap as follows.

\begin{theorem}[The log-weight gap is Gaussian]\label{thm:gap}
Fix two distinct stored windows $i$ and $j$, and let $x =\mti+\sigma_t\varepsilon$ and $\Delta_{ij}\coloneqq \mti-\mjt$. Conditional to the stored path centres $\mu_t^{(i)}$ and $\mu_t^{(j)}$, the defined log-weight gap satisfies
\[
G_{ij}:=\log\alpha_i-\log\alpha_j=\frac{\|\Delta_{ij}\|^2}{2\sigma_t^2}+\frac{\langle\Delta_{ij},\varepsilon\rangle}{\sigma_t}\ \sim\ \Norm(\kappa_{ij}/2,\ \kappa_{ij}).
\]
Therefore,  $\Prob(G_{ij}<0)=\Phi(-\tfrac12\sqrt{\kappa_{ij}})$ when $\Delta_{ij}\neq 0$.
\end{theorem}

The ratio $\kappa_{ij}$ in eq.~(\refeq{eq:kappa}) can now be identified as a \textbf{concentration parameter}: when \(\kappa_{ij}\ll1\), the two windows receive comparable weights, and when \(\kappa_{ij}\gg1\), the generating window $i$ is likely to dominate. The corresponding signal-to-noise ratio is
$
\frac{\E[G_{ij}]}{\sqrt{\Var(G_{ij})}}
=
\frac12\sqrt{\kappa_{ij}}.
$

\subsection{Replay onset and memory bank size} \label{sec:replay_onset}

We can now answer the question of when replay can be expected to occur. \cref{thm:gap} characterises the competition between windows $W_i$ and $W_j$, but trajectory replay is a statement about the full collection of $M$ posterior weights. To connect pairwise gaps to the full set of weights, fix the generating window $W_i$ and define
\[
S\coloneqq\sum_{j\neq i}e^{-G_j}\geq0,
\qquad
S_2\coloneqq\sum_{j\neq i}e^{-2G_j}\geq0.
\]
Since \(\alpha_j=\alpha_i e^{-G_j}\), we can express the $i^\text{th}$ weight as $\alpha_i=\frac{1}{1+S}$ and the effective sample size as
\[
\ESS
=
\frac1{\sum_j\alpha_j^2}
=
\frac{(1+S)^2}{1+S_2}
\leq
(1+S)^2.
\label{eq:ess_gap_relation}
\]
Thus replay corresponds to \(S\ll1\), where the aggregation of the remaining weights $\{\alpha_j\}_{j\neq i}$ is small, thus giving \(\alpha_i\approx1\) and \(\ESS\approx1\).

Let $n\coloneqq M-1$, and write
$\bar\kappa\coloneqq\frac1n\sum_{j\neq i}\kappa_{ij}$ and
$\kappa_{\min}\coloneqq\min_{j\neq i}\kappa_{ij}$ for the mean and minimum
concentration over competitors.
\begin{proposition}[Replay onset is logarithmic in $M$]
\label{prop:logM}
Under the assumptions of \cref{thm:gap} and any $\delta\in(0,1/2]$:
\begin{enumerate}
\item[(i)] \emph{(necessity)} Denote $z_\delta\coloneqq\Phi^{-1}(1-\delta)$, with probability at least
      $1-\delta$,
      \[
      \log S\ \ge\ \log n-\frac{\bar\kappa}{2}-z_\delta\sqrt{\bar\kappa}.
      \]
      In particular, if $\bar\kappa\le2(1-\eta)\log n$ for a fixed
      $\eta\in(0,1)$, then $\alpha_i\xrightarrow{\Prob}0$ as $M\to\infty$; 
\item[(ii)] \emph{(sufficiency)} if $\kappa_{\min}\ge32\log(n/\delta)$, then
      with probability at least $1-\delta$,
      \[
      1-\alpha_i\le\frac{\delta^8}{n^7},
      \qquad
      \ESS\le\Big(1+\frac{\delta^8}{n^7}\Big)^2 .
      \]
\end{enumerate}
\end{proposition}

When pairwise concentratation occurs, \cref{prop:logM} shows that replay requires $\kappa := \kappa_{\min}\approx\bar\kappa$ to grow on the order of $\log M$ as the number of competing windows increases. 

\begin{remark}[Interpretation of the replay-onset bounds]
Proposition~\ref{prop:logM} establishes the scale of the replay transition,
not a sharp threshold. The necessity and sufficiency statements involve
different summaries of the pairwise concentrations, $\bar\kappa$ and
$\kappa_{\min}$, and the constant $32$ in the sufficient condition comes
from a Gaussian tail bound followed by a union bound. When pairwise
separations themselves concentrate so that
$\kappa_{\min}$ and $\bar\kappa$ are comparable, both statements place the
transition at constant multiples of $\log M$. We therefore use
$\Theta(\log M)$ only as an onset scale, rather than identifying a universal
critical constant.
\end{remark}
 
\subsection{High dimensional scaling}

We now identify how $\kappa_{ij}$ changes with the ambient dimension $d$. 

\begin{definition}[Average squared separation of the window cloud]
\label{def:Vk_definition}
For independent windows \(W,W^{'}\) with the same law, and interpolator $\mu_t(\cdot)$, define the the average squared separation per coordinate of the interpolated window cloud as
\begin{equation}
V_{k,d}(t)
\coloneqq
\frac{1}{d}
\E\left[
\|\mu_t(W)-\mu_t(W')\|^2
\right]
=
\frac{2}{d}\tr\Cov(\mu_t(W)).
\label{eq:Vk_definition}
\end{equation}
This quantity depends on window order $k$, flow time $t$, window distribution, interpolant  $\mu_t$ and $d$.
\end{definition}

The high-dimensional replay result to be presented in \cref{thm:high_dim_replay} requires two properties of the separations in Def.~\ref{def:Vk_definition}.
First, the interpolated window cloud should retain non-vanishing spread per
coordinate as dimension increases, that is, $\liminf_{d\to\infty} V_{k,d}(t)>0.$ Second, pairwise squared separations should distribute around their population
scale $dV_{k,d}(t)$. We assume
\begin{equation}
\|\mu_t(W)-\mu_t(W')\|^2
=
dV_{k,d}(t)+O_p(\sqrt d),
\label{eq:distance_concentration}
\end{equation}
where $O_p$ denotes stochastic order in probability. Equivalently, whenever $V_{k,d}(t)$ remains bounded away from zero,
\[
\frac{
\|\mu_t(W)-\mu_t(W')\|^2
}{
dV_{k,d}(t)
}
=
1+O_p(d^{-1/2}).
\]
A sufficient condition for \cref{eq:distance_concentration} is that have finite fourth moments and sufficiently weak
cross-coordinate dependence, so that
$
\Var\!\left(
\|\mu_t(W)-\mu_t(W')\|^2
\right)
=
O(d).
$

Therefore for stored windows $W_i$ and $W_j$ whose separation obeys
\cref{eq:distance_concentration}, it follows that
\begin{equation}
\kappa_{ij}
=
\frac{\|\mu_t^{(i)}-\mu_t^{(j)}\|^2}{\sigma_t^2}
=
\frac{dV_{k,d}(t)}{\sigma_t^2}
+
O_p\!\left(
\frac{\sqrt d}{\sigma_t^2}
\right).
\label{eq:kappa_dimensional_scaling}
\end{equation}
Thus the leading concentration scale is
$dV_{k,d}(t)/\sigma_t^2$. In particular, if
$V_{k,d}(t)=\Theta(1)$ and $\sigma_t^2=\Theta(1)$, then
$\kappa_{ij}$ grows linearly with $d$ to leading order. 

\begin{theorem}[High-dimensional replay]
\label{thm:high_dim_replay}
Consider the generated query
$
x=\mu_t^{(i)}+\sigma_t\varepsilon,
\varepsilon\sim\Norm(0,I_d),
$
and assume that
$
\liminf_{d\to\infty}V_{k,d}(t)>0.
$
Assume further that the squared window separations concentrate around
$dV_{k,d}(t)$ as in \cref{eq:distance_concentration}, and that
$\sigma_t^2\le\bar\sigma^2$ for some $\bar\sigma<\infty$ independent of $d$.
For a fixed number of windows $M$, when $d\to\infty$ we obtain
$
\alpha_i(t,x)\xrightarrow{\Prob}1$ and $\ESS(t,x)\xrightarrow{\Prob}1.
$
\end{theorem}

From \cref{eq:kappa_dimensional_scaling}, when pairwise separations across windows are
comparable, we use $\kappa$ to denote their typical concentration scale:
\begin{equation}
\kappa
\asymp
\frac{dV_{k,d}(t)}{\sigma_t^2}.
\label{eq:highdimscale}
\end{equation}
Combining the order-$\log M$ replay onset from \cref{prop:logM} with
\cref{eq:highdimscale} gives
$d_\star\asymp\sigma_t^2\log M/V_{k,d}(t)$ where $\asymp$ indicates both share the same order of growth.

\begin{remark}
    Replay can thus be said to arise when the path-centre ($\mu_t$) separation grows with $d$ faster than the chosen conditional-path variance. \cref{app:compdepvariances} extends this to component-dependent variances.
\end{remark}

We include a table in \cref{tab:kappa-bandwidth} to address some bandwidths and their corresponding $\kappa$ values for reference

%% file: Chapters/Concentration.tex
\section{What controls replay concentration}
\label{chap:replayconc}

The analysis above identifies the replay concentration scale as $\kappa\asymp\frac{dV_{k,d}(t)}{\sigma_t^2}$. We use this decomposition of $\kappa$ to ask which common design choices in sequential flow matching can alter the dimensional onset of replay: coupling and interpolation, the addition of a score objective, kernel bandwidth ($\sigma_t^2$) and covariance shape.

Multi-marginal and windowed flow-matching methods replace pairwise transitions by higher order couplings and richer interpolants. We study their effects on concentration below.

\begin{proposition}[Coupling and interpolant do not alter dimensional scaling]
\label{prop:coupling_replay}
Under the assumptions of \cref{eq:distance_concentration}, for fixed
window order \(k\) with \(V_k(t)=\Theta(1)\), coupling and interpolation
affect replay concentration only through \(V_k(t)\); the leading
\(d/\sigma_t^2\) scaling in \cref{eq:kappa_dimensional_scaling} is unchanged. 
\end{proposition}

Coupling and interpolation can change \emph{what} is being replayed, but they do not change the concentration dimensional dependence directly. 

A number of sequential and multi-marginal methods augment the velocity objective with a score-matching objective. We study whether the joint score objective changes the replay mechanism.

\begin{proposition}[The score estimator inherits the replay weights]
\label{prop:score_replay}
Under the shared isotropic Gaussian conditional path
$
p_t(\cdot\mid W)=\Norm(\mu_t(W),\sigma_t^2I_d),
$
the infinite-capacity empirical minimizer $\mathcal{L}_s(s)=\E[\|s_t(X_t)-\nabla_x\log p_t(X_t\mid W)\|^2]$ of the score objective is
$
s_t^\star(x)
=
\sum_j
\alpha_j(t,x)
\frac{\mu_t^{(j)}-x}{\sigma_t^2},
$
where $\alpha_j(t,x)$ are the same posterior weights in \cref{eq:alphas}.
\end{proposition}

Consequently, adding a score objective does not change the infinite-capacity empirical concentration governing trajectory replay. We emphasise that this result does not rule out benefits from score-matching in the  finite-capacity setting.

\subsection{Bandwidth regime}

For isotropic Gaussian probability paths, the following result shows that the bandwidth $\sigma_t^2$ does alter concentration. Assume $\sigma_t^2 \asymp d^a$. Since $\kappa \asymp \frac{dV_k(t)}{d^a} = d^{1-a}V_k(t)$, the asymptotic behaviour is determined by exponent $a$ 

\begin{theorem}[Bandwidth regimes]
\label{thm:bandwidth_regimes}
Fix $M$, $k$ and $t\in(0,1)$. Assume
$V_{k,d}(t)=\Theta(1)$, pairwise separations satisfy
\cref{eq:distance_concentration}, and
$\sigma_t^2\asymp d^a$. Then, as $d\to\infty$,
\[
\begin{array}{lll}
a<1
&\Longrightarrow&
\alpha_i(t,x)\xrightarrow{\Prob}1,
\quad \ESS(t,x)\xrightarrow{\Prob}1,\\[2mm]
a=1
&\Longrightarrow&
\kappa_{ij}\text{ remains order one in probability},\\[2mm]
a>1
&\Longrightarrow&
\alpha_j(t,x)\xrightarrow{\Prob}1/M\ \forall j,
\quad \ESS(t,x)\xrightarrow{\Prob}M.
\end{array}
\]
\end{theorem}

The case $a=1$ represents neither replay nor uniform averaging. The usual dimension-independent bandwidth ($a=0$) lies in the replay regime. Linear variance scaling $\sigma_t^2=\Theta(d)$ keeps the weights non-degenerate, while faster growth drives them toward uniformity.

\subsection{Shared anisotropic covariance}

In high-dimensional nonparametric regression, structural assumptions such as anisotropic bandwidths or low-dimensional manifold structure can replace dependence on ambient dimension by a lower-dimensional quantity \citep{conn2019oracle,mcrae2020sample}. This motivates asking whether anisotropy changes the replay mechanism itself, or replaces the distance measurement from the isotropic case to an analogous anisotropic form. The following result shows that the same log-weight law holds, with Euclidean distance replaced by a covariance-weighted Mahalanobis distance.

\begin{proposition}[Anisotropic log-weight law]
\label{prop:anisotropic_gap}
Let
$
x=\mu_t^{(i)}+A\varepsilon,
\qquad
AA^\top=\Sigma_t,
\qquad
\varepsilon\sim\Norm(0,I_d),
$
and define
$
\Delta_{ij}
=
\mu_t^{(i)}-\mu_t^{(j)},
\,
\kappa_{ij}^{\Sigma}
=
\Delta_{ij}^{\top}\Sigma_t^{-1}\Delta_{ij}.
$
Conditionally on the interpolated window locations,
\[
G_{ij}
=
\log\alpha_i-\log\alpha_j
\sim
\Norm\left(
\frac{\kappa_{ij}^{\Sigma}}{2},
\kappa_{ij}^{\Sigma}
\right).
\]
\end{proposition}

Thus, anisotropy changes which stored windows are considered to be close without changing the concentration mechanism. We can therefore ask how the covariance should be shaped to reduce concentration as much as possible without increasing its total variance. 

Let
$
\Gamma
\coloneqq
\E[\Delta_{ij}\Delta_{ij}^{\top}]
$
be the covariance of the displacements between stored path centres.

\begin{proposition}[Concentration-optimal shared covariance]
\label{prop:optimal_covariance}
Assume \(\Gamma\succ0\). Among all shared covariances
\(\Sigma_t\succ0\) satisfying $\tr(\Sigma_t)=d\sigma_t^2$,
the expected concentration
$
\E[\kappa_{ij}^{\Sigma}]
=
\tr(\Sigma_t^{-1}\Gamma)
$
is minimised by
$
\Sigma_t^\star
=
\frac{d\sigma_t^2}{\tr(\Gamma^{1/2})}\Gamma^{1/2}.
$
At this optimum, $d$ is replaced by $\reff(\Gamma)$:
\[
\E[\kappa_{ij}^{\Sigma^\star}]
=
\frac{\reff(\Gamma)V_k(t)}{\sigma_t^2},
\qquad
\reff(\Gamma)
=
\frac{\bigl(\tr\Gamma^{1/2}\bigr)^2}{\tr\Gamma}.
\]
\end{proposition}

The optimal covariance therefore allocates more variance in directions where the
stored windows vary more. Under this choice, the ambient dimension \(d\) is replaced
by \(\reff(\Gamma)\), which is small when the displacement cloud is concentrated in
relatively few directions and equals \(d\) in the isotropic case.

For completeness, \cref{app:proof_elliptic} treats heavy-tailed scale-mixture
paths. These alter the rate of concentration but do not remove the high-dimensional replay limit.

\begin{remark}[Concentration is not a performance objective]
    We note here that preventing nearest-neighbour replay is not an indicator of performance improvement; sufficiently increasing the bandwidth can remove replay by replacing it with uniform averaging. We thus treat $\ESS$ as a useful description of an estimator's weighting behaviour, but not a sufficient quantity for determining performance. We defer that work to the empirical analysis.
\end{remark} 

%% file: Chapters/Coverage.tex
\section{Target-space Coverage}
\label{chap:targetspacecover}

The concentration analysis above describes how the infinite-capacity empirical estimator distributes weight across stored components, but whether the stored targets can represent the field being estimated is addressed in this section. Throughout this section, covariance is constant in flow time $\Sigma'_t=0$, so that the conditional targets $\ut(x\mid Z_j)=\dot\mu_t^{(j)}(Z_j)$ are independent of query $x$.

\begin{definition}[Target cloud and coverage functional]
\label{def:coverage}
Let $u_t^{(j)}\coloneqq\dot\mu_t(W_j)$, $j=1,\dots,M$, denote the stored
conditional targets, and let
$
\mathcal{U}_t
\coloneqq
\conv\{u_t^{(1)},\dots,u_t^{(M)}\}
\subset\R^d
$
be their convex hull. Given a reference velocity field
$f_t:\R^d\to\R^d$, representing the true target field against which the
empirical estimator is evaluated, define the target-space coverage functional
\[
D_t(x)
\coloneqq
\mathrm{dist}\!\left(f_t(x),\,\mathcal{U}_t\right)
=
\min_{w\in\mathcal{U}_t}\|f_t(x)-w\|.
\]
\end{definition}

The set $\mathcal{U}_t$ is determined by the memory bank, coupling order and interpolant. It is independent of the query \(x\) and, under the constant-covariance assumption, of the shared bandwidth or covariance. Since $v^\star_t(x)$ is a convex weighting of these targets, this produces a representation constraint, formalised in \cref{thm:coverage_floor}.

\begin{theorem}[Convex-hull representation floor]
\label{thm:coverage_floor}
Fix $t\in(0,1)$ and let $v^\star_t$ be the empirical minimizer of
\cref{lem:mainminimizer} under
$p_t(\cdot\mid W)=\Norm(\mu_t(W),\Sigma_t)$, with $\Sigma_t\succ0$ and
$\Sigma'_t=0$. Then, for every $x\in\R^d$,
\[
v^\star_t(x)\in\mathcal{U}_t,
\qquad
\|v^\star_t(x)-f_t(x)\|\geq D_t(x),
\]
where the latter bound is tight over all convex weightings of the stored targets.
Consequently, for any query law $Q$ held fixed as the shared covariance varies,
\[
\inf_{\Sigma_t\succ0}
\E_{x\sim Q}\|v^\star_t(x)-f_t(x)\|^2
\geq
\E_{x\sim Q}[D_t(x)^2].
\]
\end{theorem}

For a fixed target bank and reference field, changing the shared covariance only reweights the same stored targets and therefore leaves $D_t(x)$ unchanged. Even the optimal convex weighting cannot cross this floor. Thus, reducing replay can improve accuracy only when the target bank is already rich enough to represent the required field. A corresponding necessary condition based on the span of the target bank is given in \cref{app:proof_span_floor}. We also address a deterministic-interpolant limit in \cref{app:detinterp}. 

\begin{corollary}[Time-varying covariance extension] \label{cor:timevaryinghull}
    In practice, noise schedules can vary with time. For a shared time-varying isotropic path, writing $b_t=\dot\sigma_t/\sigma_t$ we have

    $$\mathcal{U}_t^{(b)}(x)=b_tx+\mathrm{conv}\{\dot\mu_t^{(j)}-b_t\mu_t^{(j)}\}_{j=1}^M.$$

    Consequently, Lemma 1 gives $$v^*_t(x)\in \mathcal{U}_T^{(b)}(x),\quad \|v^*_t(x)-f_t(x)\|\geq\mathrm{dist}(f_t(x),\mathcal{U}_t^{(b)}(x)).$$

    Thus, a translated, path-dependent convex-hull constraint remains. Unlike Theorem 4, the reachable set now depends on the noise schedule. Setting $b_t=0$ recovers \cref{thm:coverage_floor}. We note that $b_t$ is unchanged when the complete schedule is rescaled, i.e., $\sigma_t\mapsto s\sigma_t$; therefore, changing the bandwidth alone (the parameter controlling concentration) does not move the floor.
\end{corollary}

%% file: Chapters/Empirical.tex
\section{Empirical Validation}
\label{chap:empirical}

The theory separates two properties of the estimator: concentration (covered in \cref{chap:replay_section,chap:replayconc}) and target coverage (covered in \cref{chap:targetspacecover}). We test these two mechanisms directly on the closed-form empirical minimiser in \cref{sec:empirical-onset,sec:empirical-coverage,sec:empirical-anisotropy}, before briefly examining their transfer to finite-capacity networks in \cref{sec:empirical-neural}. Experimental details are deferred to the appendix.

\subsection{Replay onset is governed by concentration}
\label{sec:empirical-onset}

\begin{figure}[!h]
    \centering
    \includegraphics[width=0.95\linewidth]{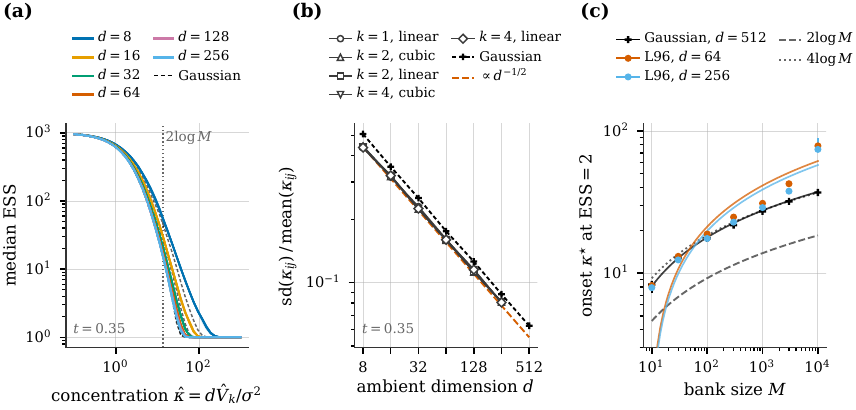}
    \caption{
    \textbf{Replay onset is controlled by the concentration parameter.}
    (a) Median effective sample size as a function of
    $\widehat\kappa=d\widehat V_k/\sigma^2$ for Lorenz--96 across ambient
    dimensions, together with an iid Gaussian control. 
    (b) Relative spread of pairwise concentrations. (c) The concentration $\kappa^\star$ at which $\ESS=2$ grows
    logarithmically with memory-bank size $M$. 
    }
    \label{fig:onset}
\end{figure}

We test whether changes in dimension, window order and interpolant are absorbed by the predicted concentration parameter $\widehat\kappa=d\widehat V_k(t)/\sigma_t^2.$

Figure~\ref{fig:onset} shows that the ESS curves collapse when expressed in terms of $\widehat\kappa$, pairwise separations concentrate at the predicted $d^{-1/2}$ relative rate, and the concentration required for replay grows logarithmically with $M$. These observations support $\kappa$, rather than ambient dimension alone, as the variable controlling the replay--smoothing transition. The measured memory-bank scaling also lies within the constant-factor bounds of \cref{prop:logM}.

\subsection{Coverage constrains the achievable error} \label{sec:empirical-coverage}

\begin{figure}[!h]
    \centering
    \includegraphics[width=0.95\linewidth]{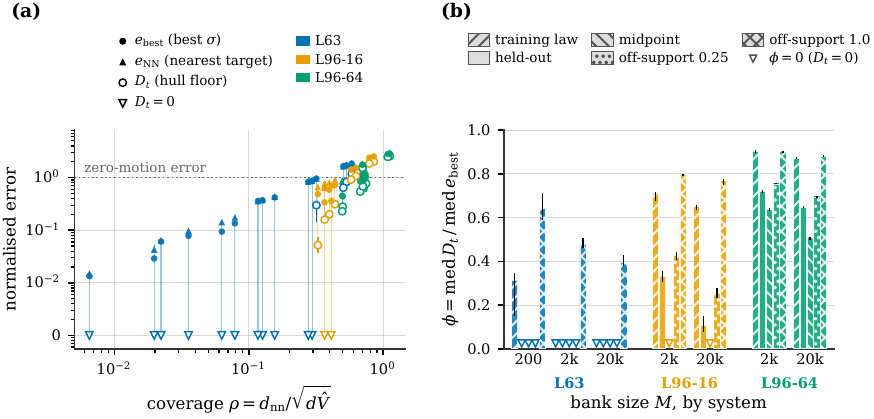}
    \caption{
    \textbf{Support coverage constrains estimator error.}
    (a) Best closed-form and nearest-neighbour error versus normalized
    query-to-bank distance $\rho$. Stems show the hull floor $D_t$;
    open markers indicate $D_t=0$, and the dashed line is the zero-motion error.
    (b) Fraction $\phi=\operatorname{med}D_t/\operatorname{med}e_{\rm best}$
    across query families and bank sizes.
    }
    \label{fig:coverage}
\end{figure}

We vary the distance of query states from the stored bank while sweeping concentration. As an observable state-space proxy for support coverage, define
$
    \rho
    =
    \frac{d_{\rm nn}}{\sqrt{d\widehat V}},
$
where $d_{\rm nn}$ is the distance from a query to its nearest stored state.
For each query family we report $e_{\rm best}$, the lowest normalized error
observed over the prescribed concentration grid. We separately compute the exact
target-space hull floor $D_t$ from \cref{thm:coverage_floor}.

Figure~\ref{fig:coverage} shows that error increases strongly as state-space coverage worsens, while a positive $D_t$ remains an uncrossable lower bound regardless of bandwidth or covariance. Thus, concentration controls how the available targets are combined, whereas coverage constrains what those targets can represent. Reducing replay can improve the weighting of a sufficiently rich target bank, but cannot remove a representation error caused by missing target directions.

\subsection{Concentration is not itself a prediction objective} \label{sec:empirical-anisotropy}

\begin{figure}[!h]
    \centering
    \includegraphics[width=\linewidth]{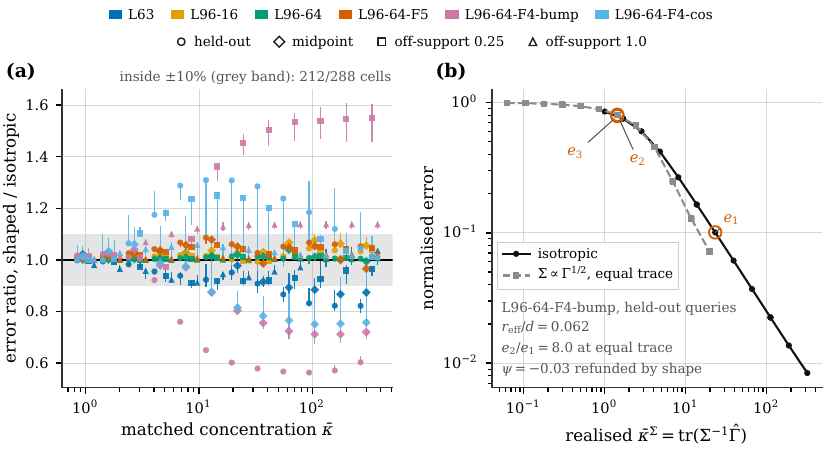}
    \caption{
    \textbf{Prediction error under matched mean concentration.}
    (a) Anisotropic/isotropic error ratio at matched realised concentration;
    212 of 288 comparisons lie within $\pm10\%$.
    (b) Equal-trace half-whitening and an isotropic kernel retuned to the
    same concentration on L96-64-F4-bump.
    }
    \label{fig:anisotropy}
\end{figure}

To examine the relationship between covariance geometry and prediction error, we compare anisotropic and isotropic kernels at realised $\bar\kappa^\Sigma=\operatorname{tr}(\Sigma^{-1}\widehat\Gamma)$; shown in Figure \ref{fig:anisotropy}. This scalar measures average Mahalanobis separation but does not uniquely determine the posterior weight distribution or effective sample size (ESS).

To assess whether the agreement in Figure \ref{fig:anisotropy}(a) reflects genuinely similar posterior concentration, we additionally compare ESS and retune the isotropic bandwidth to match the anisotropic kernel's median ESS on the same queries (Appendix \cref{table:matchcovtable}).

The additional controls show that average concentration is a reliable proxy in relatively isotropic systems, but less so in strongly anisotropic settings, where matching $\bar\kappa^\Sigma$ can yield substantially different posterior weights. ESS matching removes some apparent error differences, although residual differences remain. Thus, average concentration explains much of the observed prediction behaviour, but differences at matched $\bar\kappa^\Sigma $ cannot be attributed exclusively to covariance geometry. Proposition 5 remains an optimum for expected concentration, not prediction error.

\subsection{Finite capacity networks partially inherit replay}
\label{sec:empirical-neural}

\begin{figure}[!h]
    \centering
    \includegraphics[width=\linewidth]{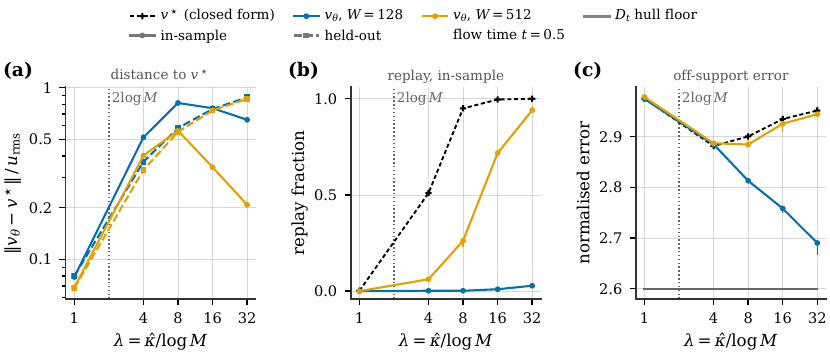}
    \vspace{0.5em}
    \caption{
    \textbf{Finite-capacity networks need not inherit the empirical minimiser.}
    (a) Network--closed-form discrepancy on in-sample and held-out queries.
    (b) In-sample single-target replay fraction.
    (c) Off-support error; the horizontal line is the hull floor $D_t$ for
    $v^\star$, and the dotted line marks $\widehat\kappa=2\log M$.
    }
    \label{fig:neuralapprox}
\end{figure}
\FloatBarrier

Lastly, we trained MLPs of widths $W\in\{128,512\}$ on a fixed Lorenz--96 problem and vary concentration relative to the $\Theta(\log M)$ replay scale. Figure~\ref{fig:neuralapprox} shows that both networks track $v^\star$ at low concentration, while the wider network inherits substantially more of its replay behaviour as concentration increases.

The narrower network departs further from the empirical minimiser and can achieve lower off-support error, where it is not constrained to the convex hull of the stored targets. An ablation study (Appendix \cref{app:empirical-details,fig:ablation}) shows that the width dependence persists across tested training budgets and weight-decay settings.

\begin{remark}[Concentration along generated trajectories]
 The preceding concentration results concern queries generated from stored conditional components. To examine their behaviour during inference, we additionally evaluate posterior ESS along trajectories generated by $v^\star $and the two trained MLPs, starting from training-law, held-out and off-support states (Appendix B, \cref{fig:essrollout}).
\end{remark}

%% file: Chapters/Appendix.tex
\section{Appendix: Proofs} \label{app:proofs}
\subsection{Proofs from Section 3}
\label{app:proofs_sec3}

\subsubsection{Proof of \cref{thm:gap}}
\label{app:proof_gap}

\begin{proof}
By the Gaussian form of the posterior weights,
\[
\alpha_j
\propto
\exp\left(
-\frac{\|x-\mjt\|^2}{2\sigma_t^2}
\right),
\]
with a normalising constant independent of $j$. Hence
\[
G_{ij}
=
\log\alpha_i-\log\alpha_j
=
\frac{
\|x-\mjt\|^2-\|x-\mti\|^2
}{
2\sigma_t^2
}.
\]
Since
\[
x-\mjt
=
\Delta_{ij}+\sigma_t\varepsilon,
\qquad
x-\mti
=
\sigma_t\varepsilon,
\]
we have
\[
\|x-\mjt\|^2
=
\|\Delta_{ij}\|^2
+
2\sigma_t\langle\Delta_{ij},\varepsilon\rangle
+
\sigma_t^2\|\varepsilon\|^2,
\]
and
\[
\|x-\mti\|^2
=
\sigma_t^2\|\varepsilon\|^2.
\]
The common quadratic noise term therefore cancels exactly, giving
\[
G_{ij}
=
\frac{\|\Delta_{ij}\|^2}{2\sigma_t^2}
+
\frac{\langle\Delta_{ij},\varepsilon\rangle}{\sigma_t}.
\]

Conditionally on $\{\mu_t^{(l)}\}$, the vector $\Delta_{ij}$ is deterministic
and $\varepsilon\sim\Norm(0,I_d)$. Therefore
\[
\langle\Delta_{ij},\varepsilon\rangle
\sim
\Norm(0,\|\Delta_{ij}\|^2),
\]
and hence
\[
\frac{\langle\Delta_{ij},\varepsilon\rangle}{\sigma_t}
\sim
\Norm(0,\kappa_{ij}).
\]
Since
\[
\frac{\|\Delta_{ij}\|^2}{2\sigma_t^2}
=
\frac{\kappa_{ij}}{2},
\]
it follows that
\[
G_{ij}
\sim
\Norm\left(
\frac{\kappa_{ij}}{2},
\kappa_{ij}
\right).
\]

For $\Delta_{ij}\neq0$, standardising gives
\[
\Prob(G_{ij}<0)
=
\Phi\left(
\frac{-\kappa_{ij}/2}{\sqrt{\kappa_{ij}}}
\right)
=
\Phi\left(
-\frac12\sqrt{\kappa_{ij}}
\right).
\]
\end{proof}

\subsubsection{Proof of \cref{prop:essbound}}
\label{app:proof_essbound}

\begin{proposition} \label{prop:essbound}
    Let
    \[
    \kappa_{\min} \coloneqq \min_{j\neq i}\kappa_{ij}.
    \]
    Under the assumptions of \cref{thm:gap}, with probability at least
    \[
    1-(M-1)e^{-\kappa_{\min}/32},
    \]
    the effective sample size satisfies
    \[
    \ESS
    \leq
    \left(
    1+(M-1)e^{-\kappa_{\min}/4}
    \right)^2.
    \]
\end{proposition}

\begin{proof}
For $j\neq i$,
\[
G_{ij}
=
\log\alpha_i-\log\alpha_j,
\]
so
\[
\alpha_j
=
\alpha_i e^{-G_{ij}}.
\]
Set $G_i\equiv0$. Using $\sum_j\alpha_j=1$,
\[
1
=
\sum_j\alpha_j
=
\alpha_i
\left(
1+\sum_{j\neq i}e^{-G_j}
\right)
=
\alpha_i(1+S),
\]
and therefore
\[
\alpha_i
=
\frac{1}{1+S}.
\]
Similarly,
\[
\sum_j\alpha_j^2
=
\alpha_i^2
\left(
1+\sum_{j\neq i}e^{-2G_j}
\right)
=
\frac{1+S_2}{(1+S)^2}.
\]
Thus
\[
\ESS
=
\frac{(1+S)^2}{1+S_2}
\leq
(1+S)^2.
\]

Fix $\theta=1/4$. By \cref{thm:gap},
\[
\Prob\left(
G_{ij}\leq\frac14\kappa_{ij}
\right)
=
\Phi\left(
-\frac14\sqrt{\kappa_{ij}}
\right).
\]
Using the Gaussian tail bound
\[
\Phi(-z)\leq e^{-z^2/2},
\qquad z>0,
\]
gives
\[
\Prob\left(
G_{ij}\leq\frac14\kappa_{ij}
\right)
\leq
e^{-\kappa_{ij}/32}
\leq
e^{-\kappa_{\min}/32}.
\]

Define the event
\[
E
\coloneqq
\left\{
G_{ij}>\frac14\kappa_{ij}
\quad
\forall j\neq i
\right\}.
\]
By a union bound,
\[
\Prob(E^c)
\leq
(M-1)e^{-\kappa_{\min}/32},
\]
so
\[
\Prob(E)
\geq
1-(M-1)e^{-\kappa_{\min}/32}.
\]

On $E$,
\[
G_{ij}
>
\frac14\kappa_{ij}
\geq
\frac14\kappa_{\min},
\]
and therefore
\[
e^{-G_j}
<
e^{-\kappa_{\min}/4}.
\]
Summing over the $M-1$ competitors,
\[
S
<
(M-1)e^{-\kappa_{\min}/4}.
\]
Substituting this into the bound on $\ESS$ yields
\[
\ESS
\leq
\left(
1+(M-1)e^{-\kappa_{\min}/4}
\right)^2
\]
on an event of probability at least
\[
1-(M-1)e^{-\kappa_{\min}/32}.
\]
\end{proof}

\subsubsection{Proof of \cref{prop:logM}}
\label{app:proof_logM}

\begin{proof}
(i) By convexity of $\exp$,
$S=\sum_{j\neq i}e^{-G_j}\ge n\exp(-\bar G)$ with
$\bar G\coloneqq\frac1n\sum_{j\neq i}G_j$. By \cref{thm:gap},
\[
\bar G=\frac{\bar\kappa}{2}+\xi,
\qquad
\xi\coloneqq\frac1n\sum_{j\neq i}\frac{\langle\Delta_{ij},\varepsilon\rangle}{\sigma_t}.
\]
Conditionally on the path centres, $\xi$ is a centred Gaussian. By the
triangle inequality in $L^2$ and Jensen's inequality,
\[
\mathrm{sd}(\xi)\le\frac1n\sum_{j\neq i}\sqrt{\kappa_{ij}}\le\sqrt{\bar\kappa}.
\]
Since $z_\delta\ge0$, $\Prob(\xi>z_\delta\sqrt{\bar\kappa})\le\Phi(-z_\delta)=\delta$,
which gives the bound on $\log S$. If $\bar\kappa\le2(1-\eta)\log n$, the
right-hand side is at least $\eta\log n-z_\delta\sqrt{2\log n}\to\infty$ for
every fixed $\delta$, so $S\xrightarrow{\Prob}\infty$ and
$\alpha_i=1/(1+S)\xrightarrow{\Prob}0$.
 
(ii) Apply \cref{prop:essbound}. Since $\kappa_{\min}\ge32\log(n/\delta)$, the
failure probability is at most $ne^{-\kappa_{\min}/32}\le\delta$. On the
complementary event,
$S<ne^{-\kappa_{\min}/4}\le n(\delta/n)^8=\delta^8/n^7$, and then
$1-\alpha_i=S/(1+S)\le S$ and $\ESS\le(1+S)^2$.
\end{proof}

\subsubsection{Proof for \cref{eq:Vk_definition}}

\begin{proof}
    Let $m\coloneqq \E[\mu_t(W)]$. For i.i.d. $\mu_t(W)$,$\mu_t(W')$
    \[
    \E\|\mu_t(W)-\mu_t(W')\|^2=\E\|(\mu_t(W)-m)-(\mu_t(W')-m)\|^2
    \]
    \[
    =\E\|\mu_t(W)-m\|^2+\E\|\mu_t(W')-m\|^2-2\langle\E[\mu_t(W)-m],\E[\mu_t(W')-m]\rangle
    \]
with the cross term factorising by independence and vanishing because each term is zero. Hence 
$$\E\|\mu_t(W)-\mu_t(W')\|^2=2\E\|\mu_t(W)-m\|^2=2\operatorname{tr}\Cov(\mu_t(W))$$
\end{proof}

\subsubsection{Proof of \cref{thm:high_dim_replay}}
\label{app:proof_highdim}

The qualitative statement of \cref{thm:high_dim_replay} uses only the
concentration assumption \cref{eq:distance_concentration}. For an explicit
rate we additionally assume a sub-Gaussian lower tail: there is $c>0$,
independent of $d$, such that for every $j\neq i$ and every
$0\le u\le\sqrt d\,V_k(t)$,
\begin{equation}
\Prob\!\left(\|\Delta_{ij}\|^2\le dV_k(t)-u\sqrt d\right)\le e^{-cu^2}.
\label{eq:appendix_distance_tail}
\end{equation}
This holds, for example, when $\|\Delta_{ij}\|^2$ is a sum of $d$ independent
nonnegative coordinate contributions with second moments bounded by $B$,
with $c=1/(2B)$ \citep{maurer2003bound}.
 
\begin{proof}
The noise $\varepsilon$ is independent of the stored windows, and
$\sigma_t^2\le\bar\sigma^2$ for all $d$.
 
Define
\[
A_d\coloneqq\Big\{\min_{j\neq i}\|\Delta_{ij}\|^2\ge\tfrac12\,dV_k(t)\Big\},
\qquad
Y_j\coloneqq\frac{\|\Delta_{ij}\|^2-dV_k(t)}{\sqrt d}.
\]
By \cref{eq:distance_concentration}, each $Y_j$ is $\mathrm{O}_p(1)$. Since $M$
is fixed, $\max_{j\neq i}|Y_j|=\mathrm{O}_p(1)$. As $V_k(t)\ge V_{\min}$,
\[
\Prob(A_d^c)
\le
\Prob\!\Big(\max_{j\neq i}|Y_j|>\tfrac12\sqrt d\,V_{\min}\Big)
\longrightarrow 0 .
\]
Under \cref{eq:appendix_distance_tail}, take $u=\tfrac12\sqrt d\,V_{\min}$.
Since $dV_k(t)-\tfrac12 dV_{\min}\ge\tfrac12 dV_k(t)$, a union bound gives
\[
\Prob(A_d^c)\le(M-1)\exp\!\big(-\tfrac{c}{4}V_{\min}^2\,d\big).
\]
 
On $A_d$,
\[
\kappa_{\min}
\ge\frac{dV_k(t)}{2\sigma_t^2}
\ge\frac{dV_{\min}}{2\bar\sigma^2}
\eqqcolon\kappa_d .
\]
Conditionally on the stored windows, \cref{prop:essbound} applies. Hence, on
$A_d$ and with conditional probability at least $1-(M-1)e^{-\kappa_d/32}$,
\[
S<s_d\coloneqq(M-1)e^{-\kappa_d/4},
\qquad
\ESS\le(1+s_d)^2,
\qquad
1-\alpha_i=\frac{S}{1+S}\le s_d .
\]
 
Let $B_d$ be the event on which the
three bounds as earlier hold. Then
\[
\Prob(B_d^c)\le\Prob(A_d^c)+(M-1)e^{-\kappa_d/32}\longrightarrow 0 .
\]
On $B_d$, $1\le\ESS\le(1+s_d)^2$ and $1-s_d\le\alpha_i\le1$. Since $s_d\to0$,
$\ESS\xrightarrow{\Prob}1$ and $\alpha_i\xrightarrow{\Prob}1$.
 
\emph{Rate.} Under \cref{eq:appendix_distance_tail}, with probability at least
\[
1-(M-1)\Big[\exp\!\big(-\tfrac c4V_{\min}^2d\big)
+\exp\!\Big(-\frac{V_{\min}d}{64\bar\sigma^2}\Big)\Big],
\]
we have
\[
0\le\ESS-1\le2s_d+s_d^2,
\qquad
0\le1-\alpha_i\le s_d,
\qquad
s_d=(M-1)\exp\!\Big(-\frac{V_{\min}d}{8\bar\sigma^2}\Big).
\]
\end{proof}

\begin{remark}[When the high-dimensional assumptions hold]
The assumption
\[
\|\Delta_{ij}\|^2=dV_{k,d}(t)+O_p(\sqrt d)
\]
requires the squared coordinate contributions to have total variance of
order $d$. A sufficient condition is uniformly bounded fourth moments
together with sufficiently weak cross-coordinate dependence, for example
when
\[
\sup_r\sum_s
\left|
\Cov(\Delta_{ij,r}^2,\Delta_{ij,s}^2)
\right|<\infty .
\]
This allows dependent coordinates, but excludes the case in which an
$O(d)$ collection of coordinates fluctuates coherently, for which the
variance of $\|\Delta_{ij}\|^2$ can instead be $O(d^2)$.

The independent windows $W,W'$ used to define $V_{k,d}(t)$ specify a
population separation scale. Rolling windows in the stored bank may overlap
and need not be independent; the theorem therefore assumes concentration
of the relevant bank separations around this scale rather than deriving it
from temporal independence. Figure~\ref{fig:onset}(b) checks this assumption
empirically for the systems used in our experiments.
\end{remark}

\subsubsection{Component-dependent variances}
\label{app:compdepvariances}

\begin{proposition}[Log-weight gap with component-dependent variances]\label{prop:hetero}
Fix $t$ and let $p_t(\cdot\mid Z_l)=\mathcal N(\mu_t^{(l)},\sigma_l^2 I_d)$ with
$\sigma_l>0$ allowed to differ across components, under equal prior weights. Let
$x=\mu_t^{(i)}+\sigma_i\varepsilon$, $\varepsilon\sim\mathcal N(0,I_d)$,
$\Delta_{ij}=\mu_t^{(i)}-\mu_t^{(j)}\neq 0$, $\kappa_{ij}=\|\Delta_{ij}\|^2/\sigma_j^2$ and
$\rho=\sigma_i/\sigma_j$. Conditionally on the path centres,
\[
G_{ij}=\mathrm{KL}(p_i\,\|\,p_j)+\rho\sqrt{\kappa_{ij}}\,\xi
+\tfrac{\rho^2-1}{2}(\xi^2-1)+\tfrac{\rho^2-1}{2}\big(Q-(d-1)\big),
\]
where $\xi\sim\mathcal N(0,1)$ and $Q\sim\chi^2_{d-1}$ are independent, and
\[
\mathrm{KL}(p_i\,\|\,p_j)=\frac{\kappa_{ij}}{2}+d\,h(\rho),\qquad
h(\rho)=\tfrac12\big(\rho^2-1-\log\rho^2\big)\ge 0,
\]
with $h(\rho)=0$ iff $\rho=1$. Moreover
$\mathrm{Var}(G_{ij})=\rho^2\kappa_{ij}+\tfrac{d}{2}(\rho^2-1)^2$.
For $\rho=1$ this reduces exactly to Theorem 1.
\end{proposition}

\begin{proof}
The normalisers no longer cancel:
$G_{ij}=d\log(\sigma_j/\sigma_i)-\frac{\|x-\mu_t^{(i)}\|^2}{2\sigma_i^2}
+\frac{\|x-\mu_t^{(j)}\|^2}{2\sigma_j^2}$.
Substituting $x-\mu_t^{(i)}=\sigma_i\varepsilon$ and
$x-\mu_t^{(j)}=\Delta_{ij}+\sigma_i\varepsilon$,
\[
G_{ij}=\frac{\|\Delta_{ij}\|^2}{2\sigma_j^2}
+\frac{\sigma_i}{\sigma_j^2}\langle\Delta_{ij},\varepsilon\rangle
+\frac{\rho^2-1}{2}\|\varepsilon\|^2-d\log\rho .
\]
Write $\xi=\langle\Delta_{ij},\varepsilon\rangle/\|\Delta_{ij}\|$ and decompose
$\|\varepsilon\|^2=\xi^2+Q$, where $Q$ is the squared norm of the component of
$\varepsilon$ orthogonal to $\Delta_{ij}$; then $\xi\sim\mathcal N(0,1)$ and
$Q\sim\chi^2_{d-1}$ are independent. Since
$\frac{\sigma_i}{\sigma_j^2}\|\Delta_{ij}\|=\rho\sqrt{\kappa_{ij}}$, collecting the
constant terms gives the representation, and the constant equals
$\mathrm{KL}(p_i\|p_j)$ by the closed form for isotropic Gaussians.
Nonnegativity of $h$ follows from $u-1-\log u\ge 0$ with $u=\rho^2$.
For the variance, $\xi$ and $\xi^2$ are uncorrelated,
$\mathrm{Var}(\xi^2)=2$ and $\mathrm{Var}(Q)=2(d-1)$.
\end{proof}

\begin{corollary}[High-dimensional replay, component-dependent variances]
Let $M$ be fixed and suppose the variance ratios $\rho_{ij}$ lie in a compact subset of
$(0,\infty)$ uniformly in $d$. If for every $j\neq i$ either
$\kappa_{ij}\ge c\,d$ with probability tending to one, or $|\log\rho_{ij}|\ge c$,
for some $c>0$ independent of $d$, then $\alpha_i(t,x)\xrightarrow{P}1$ and
$\mathrm{ESS}(t,x)\xrightarrow{P}1$ as $d\to\infty$.
\end{corollary}

\begin{proof}
Under either condition $\mathbb E[G_{ij}]\ge c' d$ for some $c'>0$, while
$\mathrm{Var}(G_{ij})=O(\kappa_{ij}+d)$. By Chebyshev,
$P\big(G_{ij}\le \tfrac12\mathbb E[G_{ij}]\big)\le 4\,\mathrm{Var}(G_{ij})/\mathbb E[G_{ij}]^2\to 0$,
so $G_{ij}\to\infty$ in probability for each of the finitely many $j\neq i$.
Hence $S=\sum_{j\neq i}e^{-G_{ij}}\to 0$ in probability, and $\alpha_i=1/(1+S)\to 1$
and $\mathrm{ESS}\le(1+S)^2\to 1$ as in \cref{prop:essbound}
\end{proof}
\

\subsection{Table relating $\kappa$ to commonly used bandwidths}\label{table:kappas}

\begin{table}[hbtp]
\centering
\label{tab:kappa-bandwidth}
\small
\begin{tabular}{llccccc}
\toprule
Path $\sigma_t^2$ & Setting & $\sigma_{1/2}^2$ & $\kappa/d$ & $d^\star$ & $\kappa\,(d{=}18)$ & $\kappa\,(d{=}137)$ \\
\midrule
$\sigma_{\min}^2 + \sigma^2 t(1-t)$ & $\sigma_{\min}{=}10^{-3},\ \sigma{=}0$    & $1.0{\times}10^{-6}$ & $2{\times}10^{6}$   & $<1$   & $3.6{\times}10^{7}$ & $2.7{\times}10^{8}$ \\
                                     & $\sigma_{\min}{=}10^{-3},\ \sigma{=}0.01$ & $2.6{\times}10^{-5}$ & $7.7{\times}10^{4}$ & $<1$   & $1.4{\times}10^{6}$ & $1.1{\times}10^{7}$ \\
                                     & $\sigma_{\min}{=}10^{-3},\ \sigma{=}0.1$  & $2.5{\times}10^{-3}$ & $800$               & $<1$   & $1.4{\times}10^{4}$ & $1.1{\times}10^{5}$ \\
$\sigma^2 t(1-t)$                    & $\sigma{=}1$                              & $0.25$               & $8$                 & $4.4$  & $144$               & $1.1{\times}10^{3}$ \\
$\sigma^2$ (constant)                & $\sigma{=}0.1$                            & $0.01$               & $200$               & $<1$   & $3.6{\times}10^{3}$ & $2.7{\times}10^{4}$ \\
                                     & $\sigma{=}1$                              & $1$                  & $2$                 & $17.5$ & $36$                & $274$ \\
$\propto d$ (Thm.~3, $a{=}1$)        & $\sigma_t^2 = c\,d$                       & $c\,d$               & $2/(cd)$            & none   & $2/c$               & $2/c$ \\
\bottomrule
\end{tabular}
\caption{Concentration $\kappa \approx d V_k/\sigma_t^2$ at commonly used bandwidths, assuming standardised coordinates and fine sampling ($V_k \approx 2$), evaluated at $t = 1/2$, where bridge-type variances peak (so $\kappa$ is smallest over flow time). $d^\star$ is the dimension at which $\kappa$ reaches the iid replay onset $4\log M$ for $M = 6240$, $4\log M = 35.0$. The rows with $\sigma_{\min} = 10^{-3}$ use the settings of \citet{lim2025paths}, which are applied in an autoencoder latent space and transfer only if that latent is approximately unit-scale. Trajectory banks replay at a larger $\kappa$ than $4\log M$ (Fig.~\ref{fig:onset}(c))}
\end{table}
\FloatBarrier

\subsection{Proofs from Section 4}
\label{app:proofs_sec4}

\subsubsection{Proof of \cref{prop:coupling_replay}}
\label{app:proof_coupling}

\begin{proof}
Under the shared isotropic Gaussian conditional path,
\[
p_t(x\mid W_j)
=
(2\pi\sigma_t^2)^{-d/2}
\exp\left(
-\frac{\|x-\mu_t^{(j)}\|^2}{2\sigma_t^2}
\right).
\]
The Gaussian prefactor is independent of $j$ and therefore cancels in the
normalised posterior weights:
\[
\alpha_j(t,x)
=
\frac{
\exp\left(
-\|x-\mu_t^{(j)}\|^2/(2\sigma_t^2)
\right)
}{
\sum_l
\exp\left(
-\|x-\mu_t^{(l)}\|^2/(2\sigma_t^2)
\right)
}.
\]
Thus the interpolant and the window order enter the weights only through the
locations $\{\mu_t^{(j)}\}$.

Taking
\[
\Delta_{ij}
=
\mu_t^{(i)}-\mu_t^{(j)},
\]
the calculation of \cref{thm:gap} applies unchanged and gives
\[
G_j
\sim
\Norm\left(
\frac{\kappa_{ij}}{2},
\kappa_{ij}
\right),
\qquad
\kappa_{ij}
=
\frac{\|\Delta_{ij}\|^2}{\sigma_t^2}.
\]

For the scaling statement, use the distance-concentration assumption
\[
\|\Delta_{ij}\|^2
=
dV_k(t)+\mathrm{O}_p(\sqrt d).
\]
Then
\[
\kappa_{ij}
=
\frac{dV_k(t)}{\sigma_t^2}
+
\mathrm{O}_p\left(
\frac{\sqrt d}{\sigma_t^2}
\right).
\]
When the window order $k$ is fixed and
\[
V_k(t)=\Theta(1),
\]
changing the window order or interpolant may change the value of $V_k(t)$,
but does not change the leading $d/\sigma_t^2$ dimensional scaling.
\end{proof}

\subsubsection{Proof of \cref{prop:score_replay}}
\label{app:proof_score}

\begin{proof}
For the shared isotropic Gaussian conditional path,
\[
p_t(x\mid W_j)
=
\Norm(\mu_t^{(j)},\sigma_t^2I_d),
\]
the conditional score is
\[
\nabla_x\log p_t(x\mid W_j)
=
\frac{\mu_t^{(j)}-x}{\sigma_t^2}.
\]

For fixed $x$, the empirical squared-loss objective is a weighted
least-squares problem over the conditional targets
\[
\frac{\mu_t^{(j)}-x}{\sigma_t^2},
\]
with weights proportional to the conditional densities
$p_t(x\mid W_j)$. By the same conditional-regression argument as
\cref{lem:mainminimizer}, its minimiser is therefore
\[
s_t^\star(x)
=
\sum_j
\alpha_j(t,x)
\frac{\mu_t^{(j)}-x}{\sigma_t^2},
\]
where
\[
\alpha_j(t,x)
=
\frac{
p_t(x\mid W_j)
}{
\sum_l p_t(x\mid W_l)
}.
\]
These are exactly the posterior responsibilities appearing in the empirical
velocity minimiser.
\end{proof}

\subsubsection{Proof of \cref{thm:bandwidth_regimes}}
\label{app:proof_bandwidth}

\begin{proof}
From \cref{eq:kappa_dimensional_scaling},
\[
\kappa_{ij}
=
\frac{dV_{k,d}(t)}{\sigma_t^2}
\left(1+O_p(d^{-1/2})\right).
\]
Since $V_{k,d}(t)=\Theta(1)$ and $\sigma_t^2\asymp d^a$,
\[
\kappa_{ij}
\asymp
d^{1-a}
\left(1+O_p(d^{-1/2})\right).
\]

If $a<1$, then $\kappa_{ij}\to\infty$ in probability. By
\cref{thm:gap},
\[
G_{ij}
=
\frac{\kappa_{ij}}{2}
+
\sqrt{\kappa_{ij}}\,Z_{ij},
\qquad
Z_{ij}\sim\Norm(0,1),
\]
and hence
\[
\frac{G_{ij}}{\kappa_{ij}}
=
\frac12
+
\frac{Z_{ij}}{\sqrt{\kappa_{ij}}}
\xrightarrow{\Prob}
\frac12.
\]
Thus $G_{ij}\to\infty$ in probability for every $j\neq i$.
Since $M$ is fixed,
\[
S=\sum_{j\neq i}e^{-G_{ij}}
\xrightarrow{\Prob}0,
\]
which gives
\[
\alpha_i=\frac{1}{1+S}\xrightarrow{\Prob}1,
\qquad
\ESS\xrightarrow{\Prob}1.
\]

If $a=1$, then $\kappa_{ij}$ remains on an order-one scale. Therefore the
mean and variance of each log-weight gap,
$\kappa_{ij}/2$ and $\kappa_{ij}$, remain order one, so increasing dimension
alone drives neither one-hot concentration nor uniform averaging.

If $a>1$, then $\kappa_{ij}\to0$ in probability. Theorem~1 therefore gives
\[
G_{ij}\xrightarrow{\Prob}0
\]
for every $j\neq i$, and hence
\[
\frac{\alpha_j}{\alpha_i}
=
e^{-G_{ij}}
\xrightarrow{\Prob}1.
\]
Since $M$ is fixed and the posterior weights sum to one,
\[
\alpha_j\xrightarrow{\Prob}\frac1M
\]
for every $j$, and consequently
\[
\ESS
=
\left(\sum_j\alpha_j^2\right)^{-1}
\xrightarrow{\Prob}M.
\]
\end{proof}

\begin{remark}[Scope of the bandwidth regimes]
The three regimes are taken with fixed memory-bank size $M$, fixed window
order $k$ and fixed flow time $t$, while the ambient dimension increases.
The assumption $V_{k,d}(t)=\Theta(1)$ rules out a simultaneous collapse of
the interpolated window cloud itself. The result should therefore not be
read as a statement about arbitrary joint growth of $M$, $k$ and $d$;
joint $M$--$d$ asymptotics would additionally require controlling the
extremes of the pairwise separation distribution.
\end{remark}

\subsubsection{Proof of \cref{prop:anisotropic_gap}}
\label{app:proof_anisotropic_gap}

\begin{proof}
The anisotropic Gaussian density is
\[
p_t(x\mid W_j)
=
(2\pi)^{-d/2}
|\Sigma_t|^{-1/2}
\exp\left(
-\frac12
(x-\mu_t^{(j)})^\top
\Sigma_t^{-1}
(x-\mu_t^{(j)})
\right).
\]
Since $\Sigma_t$ is shared across windows, the prefactor is independent of
$j$ and cancels in the posterior-weight ratio. Hence
\[
G_j
=
\frac12
\left[
(x-\mu_t^{(j)})^\top
\Sigma_t^{-1}
(x-\mu_t^{(j)})
-
(x-\mu_t^{(i)})^\top
\Sigma_t^{-1}
(x-\mu_t^{(i)})
\right].
\]

Write
\[
x
=
\mu_t^{(i)}+A\varepsilon,
\qquad
AA^\top=\Sigma_t.
\]
Then
\[
x-\mu_t^{(j)}
=
\Delta_{ij}+A\varepsilon,
\qquad
x-\mu_t^{(i)}
=
A\varepsilon.
\]
Expanding,
\[
(x-\mu_t^{(j)})^\top
\Sigma_t^{-1}
(x-\mu_t^{(j)})
=
\Delta_{ij}^\top\Sigma_t^{-1}\Delta_{ij}
+
2\Delta_{ij}^\top\Sigma_t^{-1}A\varepsilon
+
\varepsilon^\top
A^\top\Sigma_t^{-1}A
\varepsilon,
\]
whereas
\[
(x-\mu_t^{(i)})^\top
\Sigma_t^{-1}
(x-\mu_t^{(i)})
=
\varepsilon^\top
A^\top\Sigma_t^{-1}A
\varepsilon.
\]
The quadratic noise terms cancel exactly, giving
\[
G_j
=
\frac12\kappa_{ij}^{\Sigma}
+
\Delta_{ij}^\top
\Sigma_t^{-1}A\varepsilon.
\]

Conditionally on the interpolated window locations, the stochastic term is a
linear functional of $\varepsilon$. Its variance is
\[
\Delta_{ij}^\top
\Sigma_t^{-1}
AA^\top
\Sigma_t^{-1}
\Delta_{ij}
=
\Delta_{ij}^\top
\Sigma_t^{-1}
\Delta_{ij}
=
\kappa_{ij}^{\Sigma}.
\]
Therefore
\[
\Delta_{ij}^\top
\Sigma_t^{-1}A\varepsilon
\sim
\Norm(0,\kappa_{ij}^{\Sigma}),
\]
and hence
\[
G_j
\sim
\Norm\left(
\frac{\kappa_{ij}^{\Sigma}}{2},
\kappa_{ij}^{\Sigma}
\right).
\]
\end{proof}

\subsubsection{Proof of \cref{prop:optimal_covariance}}
\label{app:proof_optimal_covariance}

\begin{proof}
For a shared covariance $\Sigma_t$,
\[
\E[\kappa_{ij}^{\Sigma}]
=
\E[
\Delta_{ij}^\top
\Sigma_t^{-1}
\Delta_{ij}
]
=
\tr(\Sigma_t^{-1}\Gamma).
\]
We minimise this subject to
\[
\tr(\Sigma_t)
=
d\sigma_t^2.
\]

The Lagrangian is
\[
\mathcal{L}(\Sigma_t,\eta)
=
\tr(\Sigma_t^{-1}\Gamma)
+
\eta
\left(
\tr(\Sigma_t)-d\sigma_t^2
\right).
\]
Using
\[
\frac{\partial}{\partial\Sigma_t}
\tr(\Sigma_t^{-1}\Gamma)
=
-\Sigma_t^{-1}
\Gamma
\Sigma_t^{-1},
\]
stationarity gives
\[
-\Sigma_t^{-1}
\Gamma
\Sigma_t^{-1}
+
\eta I
=
0.
\]
Multiplying on the left and right by $\Sigma_t$,
\[
\Gamma
=
\eta\Sigma_t^2.
\]
Hence
\[
\Sigma_t
=
\eta^{-1/2}\Gamma^{1/2}.
\]
The trace constraint gives
\[
\eta^{-1/2}
=
\frac{d\sigma_t^2}
{\tr(\Gamma^{1/2})},
\]
and therefore
\[
\Sigma_t^\star
=
\frac{d\sigma_t^2}
{\tr(\Gamma^{1/2})}
\Gamma^{1/2}.
\]

Substitution gives
\[
\tr((\Sigma_t^\star)^{-1}\Gamma)
=
\frac{
\big(
\tr\Gamma^{1/2}
\big)^2
}{
d\sigma_t^2
}.
\]
Since
\[
\reff(\Gamma)
=
\frac{
(\tr\Gamma^{1/2})^2
}{
\tr\Gamma
}
\]
and
\[
\tr\Gamma
=
dV_k(t),
\]
we obtain
\[
\E[\kappa_{ij}^{\Sigma^\star}]
=
\frac{
\reff(\Gamma)V_k(t)
}{
\sigma_t^2
}.
\]

The map
\[
\Sigma_t
\mapsto
\tr(\Sigma_t^{-1}\Gamma)
\]
is convex on the positive-definite cone, so the stationary point is the
global minimum.
\end{proof}

\begin{remark}[Covariance optimisation assumptions]
The constraint $\tr(\Sigma_t)=d\sigma_t^2$ fixes the total conditional
variance, so the optimisation changes covariance shape rather than simply
increasing the bandwidth. The assumption $\Gamma\succ0$ ensures that the
closed-form optimizer is strictly positive definite. If $\Gamma$ is
singular, the same calculation applies on its support; a strictly positive
covariance on the full ambient space then requires a variance floor in the
null directions. Thus the proposition is best interpreted as a statement
about how a fixed variance budget should be allocated across directions in
which the interpolated window cloud varies.
\end{remark}

\subsubsection{Proof of the elliptic scale-mixture claim}
\label{app:proof_elliptic}

\begin{proposition}[Scale-mixture paths]
\label{prop:scale_mixture}
Let $P_\omega$ be a probability law on $(0,\infty)$ and let every window share
the conditional path
\[
p_t(x\mid W)=\int_0^\infty\Norm\!\big(x;\,\mu_t(W),\,\Sigma_t/\omega\big)\,P_\omega(d\omega),
\qquad \Sigma_t\succ0 .
\]
Generate the query as $x=\mu_t^{(i)}+\omega_0^{-1/2}A\varepsilon$ with
$AA^\top=\Sigma_t$, $\varepsilon\sim\Norm(0,I_d)$ and $\omega_0\sim P_\omega$,
independent of each other and of the stored windows. Let
$\alpha_j\propto p_t(x\mid W_j)$, $G_j=\log\alpha_i-\log\alpha_j$ and
$\kappa_{ij}^{\Sigma}=\Delta_{ij}^\top\Sigma_t^{-1}\Delta_{ij}$.
Conditionally on the interpolated window locations:
\begin{enumerate}
\item[(i)] $\Prob(G_j<0\mid\omega_0)
      =\Phi\!\big(-\tfrac12\sqrt{\omega_0\kappa_{ij}^{\Sigma}}\big)$;
\item[(ii)] $\Prob(G_j<0)\le\E\big[e^{-\omega_0\kappa_{ij}^{\Sigma}/8}\big]$.
      For the multivariate $t_\nu$ path,
      $\omega_0\sim\mathrm{Gamma}(\nu/2,\nu/2)$, this bound equals
      $\big(1+\kappa_{ij}^{\Sigma}/(4\nu)\big)^{-\nu/2}$, and
      $\Prob(G_j<0)=\Theta\big((\kappa_{ij}^{\Sigma})^{-\nu/2}\big)$
      as $\kappa_{ij}^{\Sigma}\to\infty$;
\item[(iii)] for the $t_\nu$ path with $M$ and $\nu$ fixed, if
      $\min_{j\neq i}\kappa_{ij}^{\Sigma}\ge rd$ for some $r>0$ independent
      of $d$, then $\alpha_i\xrightarrow{\Prob}1$ and
      $\ESS\xrightarrow{\Prob}1$ as $d\to\infty$.
\end{enumerate}
\end{proposition}
 
\begin{proof}
Write $Q_l\coloneqq(x-\mu_t^{(l)})^\top\Sigma_t^{-1}(x-\mu_t^{(l)})$. Because
$\Sigma_t$ is shared, $p_t(x\mid W_l)=h(Q_l)$ with
\[
h(q)=(2\pi)^{-d/2}\det(\Sigma_t)^{-1/2}
\int_0^\infty\omega^{d/2}e^{-\omega q/2}\,P_\omega(d\omega),
\]
which is strictly decreasing in $q$. Hence $G_j=\log h(Q_i)-\log h(Q_j)$, and
$G_j<0$ if and only if $Q_j<Q_i$. The weights depend on $x$ only through the
marginal density $h$, not on the realised $\omega_0$.
 
Substituting the query,
\[
Q_i=\frac{\|\varepsilon\|^2}{\omega_0},
\qquad
Q_j-Q_i=\kappa_{ij}^{\Sigma}
+2\omega_0^{-1/2}\sqrt{\kappa_{ij}^{\Sigma}}\,\eta_j,
\qquad
\eta_j\coloneqq
\frac{\Delta_{ij}^\top\Sigma_t^{-1}A\varepsilon}{\sqrt{\kappa_{ij}^{\Sigma}}}
\sim\Norm(0,1),
\]
using $\Var(\Delta_{ij}^\top\Sigma_t^{-1}A\varepsilon)=\kappa_{ij}^{\Sigma}$ as
in \cref{prop:anisotropic_gap}.
 
\emph{(i).}
$\Prob(Q_j<Q_i\mid\omega_0)
=\Prob\big(\eta_j<-\tfrac12\sqrt{\omega_0\kappa_{ij}^{\Sigma}}\big)
=\Phi\big(-\tfrac12\sqrt{\omega_0\kappa_{ij}^{\Sigma}}\big)$.
 
\emph{(ii).} From $\Phi(-z)\le e^{-z^2/2}$,
$\Prob(G_j<0\mid\omega_0)\le e^{-\omega_0\kappa_{ij}^{\Sigma}/8}$; take
expectations over $\omega_0$. For $\omega_0\sim\mathrm{Gamma}(\nu/2,\nu/2)$,
$\E[e^{-s\omega_0}]=(1+2s/\nu)^{-\nu/2}$; set $s=\kappa_{ij}^{\Sigma}/8$.
For the matching lower bound, write $\kappa=\kappa_{ij}^{\Sigma}\ge1$. If
$\omega_0\le1/\kappa$ then
$\Phi(-\tfrac12\sqrt{\omega_0\kappa})\ge\Phi(-\tfrac12)$. The Gamma density
is at least $c_\nu\,\omega^{\nu/2-1}$ on $(0,1]$ with
$c_\nu=(\nu/2)^{\nu/2}e^{-\nu/2}/\Gamma(\nu/2)$, so
\[
\Prob(G_j<0)
\ge\Phi(-\tfrac12)\,\Prob(\omega_0\le1/\kappa)
\ge\Phi(-\tfrac12)\,\frac{2c_\nu}{\nu}\,\kappa^{-\nu/2}.
\]
 
\emph{(iii).} For the $t_\nu$ path, $h(q)\propto(1+q/\nu)^{-(\nu+d)/2}$, so
\[
G_j=\frac{\nu+d}{2}\log\!\Big(1+\frac{Q_j-Q_i}{\nu+Q_i}\Big).
\]
Fix $\omega_0>0$ and let
\[
E\coloneqq
\Big\{\eta_j\ge-\tfrac14\sqrt{\omega_0\kappa_{ij}^{\Sigma}}\ \ \forall j\neq i\Big\}
\cap\big\{\|\varepsilon\|^2\le2d\big\}.
\]
By $\Phi(-z)\le e^{-z^2/2}$, a union bound, and the chi-square tail
$\Prob(\|\varepsilon\|^2\ge d+2\sqrt{dx}+2x)\le e^{-x}$ at $x=d/8$
\citep{laurent2000adaptive},
\[
\Prob(E^c\mid\omega_0)\le(M-1)e^{-\omega_0 rd/32}+e^{-d/8}.
\]
On $E$, $Q_j-Q_i\ge\kappa_{ij}^{\Sigma}/2\ge rd/2$ and $Q_i\le2d/\omega_0$.
For $d\ge\nu\omega_0$,
\[
\frac{Q_j-Q_i}{\nu+Q_i}
\ge\frac{r\omega_0 d}{2(\nu\omega_0+2d)}
\ge\frac{r\omega_0}{6},
\qquad\text{so}\qquad
G_j\ge\frac d2\log\!\Big(1+\frac{r\omega_0}{6}\Big).
\]
Therefore, on $E$,
$S=\sum_{j\neq i}e^{-G_j}\le(M-1)(1+r\omega_0/6)^{-d/2}\to0$, giving
$\alpha_i=1/(1+S)\to1$ and $\ESS\le(1+S)^2\to1$ with conditional probability
tending to one for every fixed $\omega_0>0$. Since $\omega_0>0$ almost surely,
dominated convergence over $\omega_0$ gives the unconditional statement.
\end{proof}
 
\begin{remark}
Taking $P_\omega=\delta_1$ in (i) recovers the Gaussian misranking probability
$\Phi(-\tfrac12\sqrt{\kappa_{ij}})$ of \cref{thm:gap}. Heavy tails therefore do
not change the direction of the high-dimensional limit; they convert
exponential misranking decay into polynomial decay. Under
$\kappa_{ij}^{\Sigma}=\Theta(d)$, a fraction $\Theta(d^{-\nu/2})$ of queries
avoids concentration on the generating window.
\end{remark}

\subsection{Proofs from Section 5}
\label{app:proofs_sec5}

\subsubsection{Proof of \cref{thm:coverage_floor}}
\label{app:proof_coverage_floor}

For this result we require the shared covariance to be constant in flow time,
so that the conditional velocity targets are
\[
u_t^{(j)}
=
\dot\mu_t(W_j)
\]
and contain no covariance-derivative term. In the isotropic case this is the
condition $\sigma_t'=0$; for a general shared covariance it corresponds to
$\dot\Sigma_t=0$.

\begin{proof}
By \cref{lem:mainminimizer},
\[
v_t^\star(x)
=
\sum_j
\alpha_j(t,x)
u_t^{(j)},
\]
where
\[
\alpha_j(t,x)
=
\frac{
p_t(x\mid W_j)
}{
\sum_l p_t(x\mid W_l)
}.
\]
Since $\Sigma_t\succ0$, every Gaussian density is strictly positive, so
\[
\alpha_j(t,x)\geq0,
\qquad
\sum_j\alpha_j(t,x)=1.
\]
Hence
\[
v_t^\star(x)
\in
\conv\{
u_t^{(1)},\ldots,u_t^{(M)}
\}
=
\mathcal{U}_t,
\]
proving convex-hull inclusion.

By definition,
\[
D_t(x)
=
\min_{w\in\mathcal{U}_t}
\|f_t(x)-w\|.
\]
Since
\[
v_t^\star(x)\in\mathcal{U}_t,
\]
it follows immediately that
\[
\|v_t^\star(x)-f_t(x)\|
\geq
D_t(x),
\]
proving the pointwise lower bound.

For the infimum result, the map
\[
\alpha
\mapsto
\sum_j\alpha_j u_t^{(j)}
\]
maps the simplex
\[
\mathcal{S}_{M-1}
=
\left\{
\alpha\in\R_{\geq0}^M:
\sum_j\alpha_j=1
\right\}
\]
onto the convex hull $\mathcal{U}_t$. Therefore
\[
\left\{
\sum_j\alpha_j u_t^{(j)}
:
\alpha\in\mathcal{S}_{M-1}
\right\}
=
\mathcal{U}_t,
\]
and hence
\[
\inf_{\alpha\in\mathcal{S}_{M-1}}
\left\|
f_t(x)
-
\sum_j\alpha_j u_t^{(j)}
\right\|
=
\min_{w\in\mathcal{U}_t}
\|f_t(x)-w\|
=
D_t(x).
\]

Additionally, under the constant-in-flow-time covariance assumption,
\[
u_t^{(j)}
=
\dot\mu_t(W_j).
\]
Thus the target cloud
\[
\mathcal{U}_t
=
\conv\{
u_t^{(1)},\ldots,u_t^{(M)}
\}
\]
depends on the memory bank and interpolant but not on the shared covariance.
Changing $\Sigma_t$ changes the posterior weights but not
$\mathcal{U}_t$, and therefore does not change $D_t(x)$.

Finally, fix a query law $Q$ independently of the covariance being varied.
From the pointwise lower bound,
\[
\|v_t^\star(x)-f_t(x)\|^2
\geq
D_t(x)^2.
\]
Taking expectation under $Q$ gives
\[
\E_{x\sim Q}
\|v_t^\star(x)-f_t(x)\|^2
\geq
\E_{x\sim Q}
D_t(x)^2.
\]
Since the right-hand side is independent of $\Sigma_t$,
\[
\inf_{\Sigma_t\succ0}
\E_{x\sim Q}
\|v_t^\star(x)-f_t(x)\|^2
\geq
\E_{x\sim Q}
D_t(x)^2.
\]
\end{proof}

\begin{remark}[Why constant covariance is required]
The constant-in-flow-time covariance assumption is essential for the fixed
convex-hull statement. When $\dot\Sigma_t=0$, the conditional targets are
$u_t^{(j)}=\dot\mu_t(W_j)$ and changing the shared covariance changes only
the posterior weights, not the set of available targets. If
$\dot\Sigma_t\neq0$, the conditional velocity contains an additional
covariance-derivative term depending on both $x$ and the path covariance,
so changing the covariance also changes the targets themselves. In that
case the fixed target cloud $\mathcal U_t$ used in Theorem~\ref{thm:coverage_floor}
is no longer invariant to covariance choice. We refer to the \cref{cor:timevaryinghull} proof given later for the time varying case.
\end{remark}

\subsubsection{Target Span Deficiency and Proof}

\begin{corollary}[Target-span deficiency]
\label{cor:span_floor}
Let
$
T_t\coloneqq\mathrm{span}\{u_t^{(j)}-u_t^{(1)}:j=2,\dots,M\}
$
and let $\Pperp$ denote orthogonal projection onto $T_t^{\perp}$. Then for
every $x$,
\[
D_t(x)
\ \geq\
\big\|\Pperp\big(f_t(x)-u_t^{(1)}\big)\big\|.
\]
In particular, if \(\dim T_t<d\) and \(f_t(x)-u_t^{(1)}\) has a nonzero component
orthogonal to \(T_t\), the floor is strictly positive regardless of the
bandwidth, shared covariance, or choice of convex weights.
\end{corollary}

\label{app:proof_span_floor}

\begin{proof}
For any
\[
\alpha
\in
\mathcal{S}_{M-1},
\]
we have
\[
\sum_j
\alpha_j u_t^{(j)}
-
u_t^{(1)}
=
\sum_j
\alpha_j
\left(
u_t^{(j)}-u_t^{(1)}
\right)
\in
T_t,
\]
using
\[
\sum_j\alpha_j=1.
\]
Therefore
\[
\Pperp
\left(
\sum_j
\alpha_j u_t^{(j)}
-
u_t^{(1)}
\right)
=
0.
\]
It follows that
\[
\Pperp
\left(
f_t(x)
-
\sum_j
\alpha_j u_t^{(j)}
\right)
=
\Pperp
\left(
f_t(x)-u_t^{(1)}
\right).
\]

Since $\Pperp$ is an orthogonal projection,
\[
\|\Pperp z\|
\leq
\|z\|
\]
for every $z\in\R^d$. Thus
\[
\left\|
f_t(x)
-
\sum_j
\alpha_j u_t^{(j)}
\right\|
\geq
\left\|
\Pperp
\left(
f_t(x)-u_t^{(1)}
\right)
\right\|
\]
for every
\[
\alpha
\in
\mathcal{S}_{M-1}.
\]
Taking the infimum over $\alpha$ and taking the infimum over all convex weights as in
\cref{thm:coverage_floor} gives
\[
D_t(x)
\geq
\left\|
\Pperp
\left(
f_t(x)-u_t^{(1)}
\right)
\right\|.
\]
\end{proof}

\subsection{Deterministic-interpolant limit and proof}
\label{app:detinterp}

\begin{corollary}[Deterministic-interpolant limit]\label{cor:det}
Fix $t\in(0,1)$ and a shared isotropic path $p_t(\cdot\mid Z)=\mathcal N(\mu_t(Z),\sigma^2 I_d)$
with $\sigma$ constant in flow time, and let the empirical coupling place weights
$\pi_j>0$ on conditioning variables $Z_j$. For every $x$ with a unique nearest centre
$i(x)=\arg\min_j\|x-\mu_t^{(j)}\|$, which holds for Lebesgue-almost every $x$,
\[
\lim_{\sigma\downarrow 0}\alpha_j(t,x)=\mathbb 1\{j=i(x)\},\qquad
\lim_{\sigma\downarrow 0}v_t^\star(x)=u_t^{(i(x))}.
\]
Consequently the limiting field equals a single stored target, and
\[
\big\|v_t^\star(x)-f_t(x)\big\|\;\to\;\big\|u_t^{(i(x))}-f_t(x)\big\|
\;\ge\;\min_j\big\|u_t^{(j)}-f_t(x)\big\|\;\ge\;D_t(x).
\]
\end{corollary}

\paragraph{Proof of Corollary~\ref{cor:det}.}
With weights $\pi_j$, Lemma 1 generalises to
$\alpha_j\propto \pi_j\,p_t(x\mid Z_j)$. For $j\neq i:=i(x)$,
\[
\log\alpha_j-\log\alpha_i=\log\frac{\pi_j}{\pi_i}
-\frac{\|x-\mu_t^{(j)}\|^2-\|x-\mu_t^{(i)}\|^2}{2\sigma^2}.
\]
The first term is finite and independent of $\sigma$; the second tends to $-\infty$
because the bracket is strictly positive by uniqueness of the nearest centre.
Hence $\alpha_j\to 0$ for $j\neq i$ and $\alpha_i\to 1$. Since $\sigma$ is constant in
flow time, $u_t^{(j)}=\dot\mu_t(Z_j)$ is independent of $x$ and $\sigma$, so
$v_t^\star(x)=\sum_j\alpha_j u_t^{(j)}\to u_t^{(i)}$. The first inequality is immediate;
the second holds because every stored target lies in $\mathcal U_t$.
Points with a non-unique nearest centre lie on finitely many hyperplanes and have
Lebesgue measure zero.

Some multi-marginal methods train the vector field on deterministic interpolants $\sigma_t\equiv 0$, particularly \cite{kviman2026} whose learned interpolant is regressed without added noise in its released implementation. For this case, the conditional paths have no density and Lemma 1 does not apply directly: the population minimiser is determined only on the finite set $\{\mu_t^{(j)}\}$ and is arbitrary everywhere else. \cref{cor:det} identifies the zero-bandwidth limit of the Gaussian-smoothed empirical minimiser. It does not uniquely determine the off-support extension learned by a model trained directly with deterministic interpolants.

\subsection{Proof of \cref{cor:timevaryinghull}}
\begin{proof}
    For a shared time-varying isotropic path, writing $b_t=\dot\sigma_t/\sigma_t$ we have

    $$u_t^{(j)}(x)=\dot\mu_t^{(j)}+b_t(x-\mu_t^{(j)}).$$

    We can write $\mathcal{U}_t^{(b)}(x)=b_tx+\mathrm{conv}\{\dot\mu_t^{(j)}-b_t\mu_t^{(j)}:j=1,\dots,M\}$

    Consequently, Lemma 1 gives 

    $$v^*_t(x)\in \mathcal{U}_t^{(b)}(x).$$
    Since $\mathcal U_t^{(b)}(x)$ is compact,
    \[
    \|v_t^\star(x)-f_t(x)\|\;\ge\;\min_{w\in\mathcal U_t^{(b)}(x)}\|f_t(x)-w\|=\mathrm{dist}\big(f_t(x),\mathcal U_t^{(b)}(x)\big).
    \]
    As in Theorem~4, the map $\alpha\mapsto b_t x+\sum_j\alpha_j y_t^{(j)}$ sends the simplex $\mathcal S_{M-1}$ onto $\mathcal U_t^{(b)}(x)$, so the bound is tight over all convex weightings. By translation invariance of the distance, $\mathrm{dist}\big(f_t(x),\mathcal U_t^{(b)}(x)\big)=\mathrm{dist}\big(f_t(x)-b_t x,\ \mathrm{conv}\{y_t^{(j)}\}\big)$.
    
    For the rescaling, let $\tilde\sigma_t=s\,\sigma_t$ with constant $s>0$. Then $\tilde b_t=s\dot\sigma_t/(s\sigma_t)=b_t$. The centres $\mu_t^{(j)}$ and their derivatives do not depend on $\sigma_t$, so the set $\mathcal U_t^{(b)}(x)$ and the floor are unchanged. Rescaling changes only the weights $\alpha_j$, and therefore only where $v_t^\star(x)$ sits inside the same set.
    
\end{proof}

\section{Appendix: Empirical Details}
\label{app:empirical-details}

\subsection{Common closed-form protocol.}
The experiments in \cref{sec:empirical-onset,sec:empirical-coverage,sec:empirical-anisotropy} evaluate the infinite-capacity empirical flow-matching estimator directly;
no network is trained. Posterior weights and the corresponding velocity
field are evaluated in closed form. For the dynamical systems, ground-truth
targets are exact one-step displacements obtained by numerically integrating
the underlying ODE.

Unless otherwise stated, every experimental cell contains $500$ evaluation
queries and is repeated over five seeds, $0,\ldots,4$. Random draws are
deterministically keyed by system, seed and stream, so results are independent
of execution order. Banks of different sizes are nested subsets of a common
random permutation within each system and seed. Query families receive their
own deterministic streams and are therefore identical when reused across
bank sizes, concentration values or covariance geometries.

Within each seed we first take the median over evaluation queries. Across
seeds we report the median of these seed-level values, with the seed
minimum--maximum as the error bar. Paired quantities, including error ratios
and differences, are formed within each seed before aggregation. All
concentration grids and matching tolerances are specified in the experiment
configuration before evaluation.

\subsection{Dynamical systems.}
We use Lorenz--63 with the standard parameters
$(\sigma,\rho,\beta)=(10,28,8/3)$, integrated with step size $0.005$ and
recorded every $10$ integration steps, and Lorenz--96
\[
\dot x_i=(x_{i+1}-x_{i-2})x_{i-1}-x_i+F
\]
with integration step $0.01$ and observations every $5$ integration steps.
Thus all dynamical-system experiments use observation interval
$\Delta\tau=0.05$.

The standard Lorenz--96 systems use $F=8$. To probe covariance geometries
with different displacement spectra, the anisotropy experiment additionally
uses a $d=64$, $F=5$ system and two $d=64$, $F=4$ systems. The latter are
initialised in two different structured regimes: \texttt{F4-bump} starts
from the constant forcing state with a $0.01$ perturbation at one coordinate,
whereas \texttt{F4-cos} starts from
\[
x_i(0)=F+0.5\cos(2\pi i/d).
\]
These variants use the same Lorenz--96 equations and numerical integration;
only the forcing and initial condition differ.

The replay-onset experiment additionally includes iid
$\Norm(0,I_d)$ banks. These have no dynamics or prediction target and are
used solely as a control whose pairwise-separation geometry is known
independently of an attractor.

\subsection{Replay onset and dimensional concentration.}
For \cref{fig:onset}, we test three predictions: whether weight behaviour
collapses onto the concentration parameter
$\widehat\kappa=d\widehat V_k/\sigma_t^2$ across dimension, window order and
interpolant; whether pairwise separations concentrate at relative rate
$d^{-1/2}$; and whether the concentration required for replay grows
logarithmically with memory-bank size.

For each interpolated window cloud we compute
\[
\widehat V_k(t)
=
\frac{2}{d}\tr\widehat{\Cov}(\mu_t(W)),
\qquad
\widehat\kappa
=
\frac{d\widehat V_k(t)}{\sigma_t^2},
\]
and evaluate the exact posterior weights. Weight spread is summarized by
\[
\ESS
=
\left(\sum_j\alpha_j^2\right)^{-1}.
\]

The principal sweep uses Lorenz--96 dimensions
$d\in\{8,16,32,64,128,256\}$, window orders $k\in\{1,2,4\}$,
and $M=1000$ stored windows. We evaluate flow times
$t\in\{0.5,0.35\}$. The second value is included because $t=0.5$ is an
interpolation knot for the even window orders, at which the linear and cubic
interpolants coincide; $t=0.35$ therefore provides a nontrivial comparison
of interpolation rules. The concentration sweep contains $40$ values over
$\widehat\kappa\in[0.1,1000]$.

The same measurements are repeated on iid Gaussian banks at
$d\in\{8,16,32,64,128,256,512\}$. In \cref{fig:onset}(b) we report
\[
\frac{\operatorname{sd}(\kappa_{ij})}
     {\operatorname{mean}(\kappa_{ij})}
\]
and compare it with the $d^{-1/2}$ relative-concentration prediction.

For the memory-bank sweep,
$M\in\{10,30,100,300,1000,3000,10000\}$ on
\texttt{Gauss-512}, \texttt{L96-64-onset}, and
\texttt{L96-256-onset}. Banks are nested across $M$. For each bank size,
the empirical replay onset $\kappa^\star$ is defined as the concentration
at which the median effective sample size reaches $\ESS=2$.

\subsection{Coverage and attainable prediction error.}
For \cref{fig:coverage}, we vary the distance of a query from the stored
state bank while independently sweeping kernel concentration. Let
\[
d_{\rm nn}(x)=\min_j\|x-X_j\|
\]
be the Euclidean distance to the nearest stored state. We normalize this by
the typical bank separation,
\[
\rho(x)
=
\frac{d_{\rm nn}(x)}{\sqrt{d\widehat V}},
\]
and use $\rho$ as an observable state-space coverage proxy.

Five query families are used. \texttt{heldout} consists of states sampled
without replacement from an independently simulated test trajectory.
\texttt{isotropic-noise} samples a stored centre and adds independent
$\Norm(0,I_d)$ noise, corresponding to the Gaussian flow-matching training
query law. The two off-support families start from a held-out state and add
a perturbation in a uniformly random direction of norm
\[
0.25\sqrt{d\widehat V}
\quad\text{or}\quad
1.0\sqrt{d\widehat V},
\]
respectively. Finally, \texttt{midpoint} takes the midpoint of two
independently sampled stored centres. Each named family is generated once
per system and seed and then reused across all bank sizes and concentration
values.

The experiment uses Lorenz--63 with
$M\in\{200,2000,20000\}$ and Lorenz--96 at
$d\in\{16,64\}$ with $M\in\{2000,20000\}$. For each bank,
concentration is swept over $20$ values beginning at $\kappa=0.3$ and
extending to $64\times2\log M$, so all memory-bank sizes are explored to
the same depth relative to their predicted replay onset.

For each family we report $e_{\rm best}$, the lowest normalized closed-form
prediction error observed over this pre-specified concentration grid, and
the nearest-neighbour error $e_{\rm NN}$. We separately compute the
target-space representation floor
\[
D_t(x)
=
\operatorname{dist}\!\left(f_t(x),\mathcal U_t\right),
\]
by projecting the reference target onto the convex hull of stored targets
using non-negative least squares with an explicit simplex constraint.
Errors are normalized by the median magnitude of the true one-step field on
held-out trajectory states.

\subsection{Matched-concentration covariance experiment.}
For \cref{fig:anisotropy}, we separate the effect of covariance shape from
the effect of changing overall posterior concentration. For a shared
covariance $\Sigma$, realised concentration is summarized by
\[
\bar\kappa^\Sigma
=
\tr(\Sigma^{-1}\widehat\Gamma),
\]
where $\widehat\Gamma$ is the empirical covariance of pairwise interpolated
centre displacements.

All experiments use $M=20000$ and the query families
\texttt{heldout}, \texttt{off-0.25}, \texttt{off-1.0}, and
\texttt{midpoint}. We evaluate Lorenz--63, standard Lorenz--96 at
$d=16$ and $d=64$, and the three controlled $d=64$ Lorenz--96 variants
\texttt{F5}, \texttt{F4-bump}, and \texttt{F4-cos}. The concentration grid
contains $12$ operating points.

For \cref{fig:anisotropy}(a), isotropic and anisotropic kernels are compared
only when their realised concentrations agree within the pre-specified
relative tolerance of $10\%$. The same stored bank and query family are
used on both sides of every comparison, and the error ratio is formed within
seed before aggregation. 

In \cref{fig:anisotropy}(b), we additionally compare the equal-trace
concentration-optimal covariance
\[
\Sigma\propto\widehat\Gamma^{1/2}
\]
with an isotropic covariance under the same total variance budget. The
isotropic bandwidth is then independently retuned to match the realised
concentration of the anisotropic kernel. 

\subsection{Matched ESS covariance control}To determine whether matching average concentration also matches the posterior weight distribution, we evaluate posterior ESS under the isotropic and anisotropic kernels using the same stored banks and query sets as Figure 4.

For each anisotropic operating point, we additionally select an isotropic bandwidth that matches its median ESS on the same queries. We then repeat the paired prediction-error comparison using the original ±10\% error-ratio criterion. A comparison is counted as lying within the tolerance when its entire observed seed range is contained within [0.9, 1.1]. Operating points where the anisotropic ESS is saturated or cannot be bracketed by the isotropic bandwidth sweep are excluded from the ESS-matched comparison.

\begin{table}[htbp]
    \centering
    \label{table:matchcovtable}
    \resizebox{\textwidth}{!}{%
        \begin{tabular}{lcccc}
        System & $r_{\rm eff}/d$ & median $|\log(\mathrm{ESS}_\Sigma/\mathrm{ESS}_{\rm iso})|$ & within $\pm10\%$, matched $\bar\kappa$ & within $\pm10\%$, matched ESS \\
        \hline
        L96-64 & 0.92 & 0.08 & 48/48 & 48/48 \\
        L96-16 & 0.92 & 0.14 & 47/48 & 48/48 \\
        L96-64-F5 & 0.76 & 0.22 & 48/48 & 48/48 \\
        L63 & 0.82 & 0.08 & 33/48 & 39/47 \\
        L96-64-F4-cos & 0.32 & 1.69 & 19/48 & 29/48 \\
        L96-64-F4-bump & 0.06 & 0.38 & 17/48 & 17/29 \\
        \end{tabular}%
    }
    \caption{Prediction-error agreement under matched concentration and ESS}
\end{table}

On the relatively isotropic Lorenz-96 systems, both matching criteria give almost identical results. On strongly anisotropic systems, however, matched mean concentration need not imply similar ESS. In particular, the count for L96-64-F4-cos increases from 19/48 to 29/48 when matching ESS instead of $\bar\kappa^\Sigma$.

These results indicate that some differences arise from unmatched posterior concentration. Nevertheless, residual prediction-error differences remain even after ESS matching. Since ESS also does not uniquely determine the full posterior weight distribution, these residuals cannot be attributed exclusively to covariance shape.

\subsection{Finite-capacity experiment.}
For \cref{fig:neuralapprox}, we use Lorenz--96 with $d=64$, $F=8$,
$\Delta\tau=0.05$ and a fixed bank of $M=2000$ pairwise transitions.
We vary
\[
\lambda=\frac{\widehat\kappa}{\log M}
\in\{1,4,8,16,32\},
\]
and train velocity MLPs of widths $W\in\{128,512\}$ for $50{,}000$ steps
over three seeds. The network input is
$[x,\operatorname{emb}(t)]$, where
$\operatorname{emb}(t)$ contains sine and cosine Fourier features with
frequencies $1,\ldots,16$; the network has three hidden SiLU layers and a
linear $d$-dimensional output, with no normalization or residual layers.

Each training batch samples a stored transition $j$, flow time
$t\sim\mathrm{Unif}(0,1)$ and $\varepsilon\sim\Norm(0,I_d)$, forms
\[
x=(1-t)X_0^{(j)}+tX_1^{(j)}+\sigma\varepsilon,
\]
and minimizes $\|v_\theta(x,t)-u_j\|^2$. We use Adam with batch size
$512$, $\beta=(0.9,0.999)$ and no weight decay. The learning rate warms
linearly for $500$ steps to $10^{-3}$ and then follows cosine decay to
$10^{-5}$; gradients are clipped to norm $1$. The same hyperparameters
are used for every $(\lambda,W)$ cell, with no early stopping, EMA or
augmentation.

The memory bank, held-out data, evaluation queries and closed-form targets
are fixed across training seeds, so seed variation reflects training
randomness alone. We evaluate at
$t\in\{0.25,0.5,0.75\}$ and checkpoints
$1{,}000$, $3{,}000$, $10{,}000$, $30{,}000$ and $50{,}000$; the main
figure reports the final checkpoint at $t=0.5$. Each cell contains $500$
evaluation queries. We report discrepancy from the closed-form minimizer,
single-target replay fraction and prediction error on held-out and
off-support queries.

\begin{figure}[!h]
    \centering
    \includegraphics[width=0.8\linewidth]{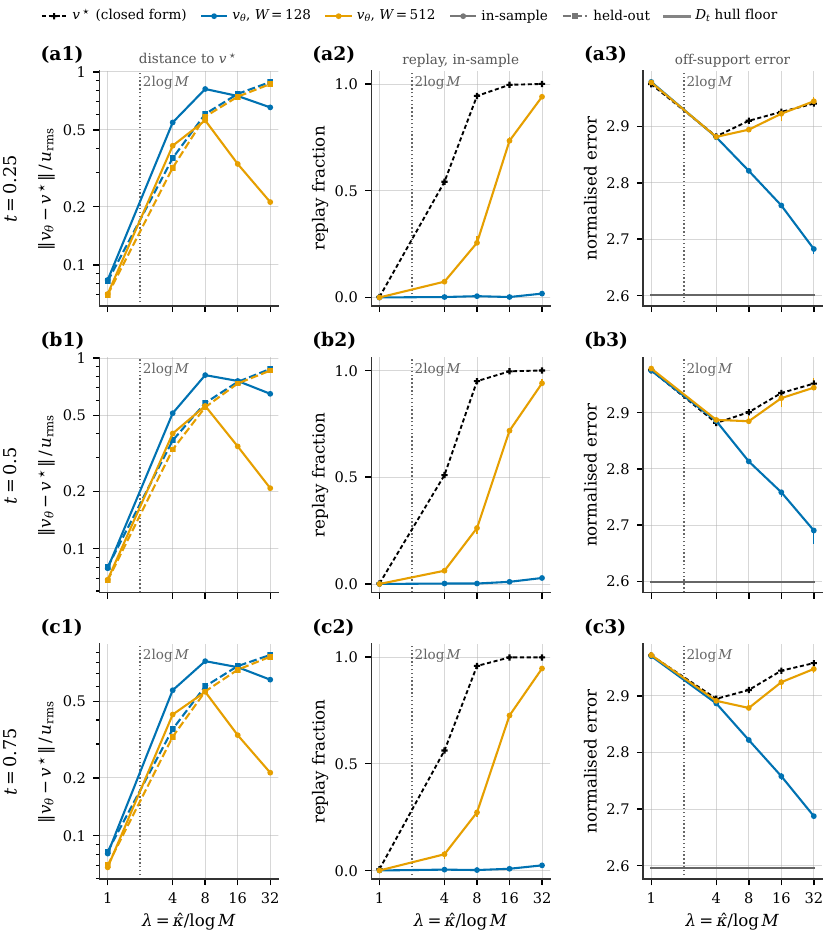}
    \caption{
    \textbf{Finite-capacity replay behaviour is consistent across flow time.}
    Rows correspond to $t\in\{0.25,0.5,0.75\}$.
    The first column shows network--closed-form discrepancy on in-sample
    (solid) and held-out (dashed) queries; the second shows the in-sample
    single-target replay fraction; and the third shows off-support prediction
    error for the \texttt{off-1.0} query family.
    The wider network increasingly inherits the replay behaviour of $v^\star$
    as concentration grows, whereas the narrower network remains further from
    the empirical minimiser and can attain lower off-support error.
    The vertical dotted line marks $\widehat\kappa=2\log M$.
    }
    \label{fig:flowtimeconsistency}
\end{figure}

\begin{figure}[!h]
    \centering
    \begin{subfigure}[b]{0.8\textwidth}
        \centering
        \includegraphics[width=\textwidth]{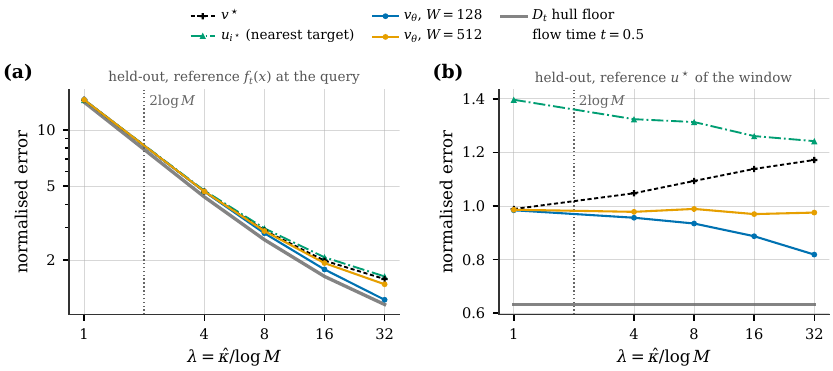}
        \caption{
        \textbf{Held-out prediction error under two reference fields.}
        At $t=0.5$, we compare the closed-form field $v^\star$, the trained networks,
        and the nearest stored target across concentration
        $\lambda=\widehat\kappa/\log M$.
        (\romannumeral 1) Error is measured against the exact true field $f_t(x)$ at the query;
        because the held-out query contains path noise whose scale varies with
        $\lambda$, the corresponding hull floor $D_t$ also varies.
        (\romannumeral 2) Error is instead measured against the held-out window's own displacement
        $u^\star$, for which the target bank and hull floor are fixed across
        $\lambda$. This separates approximation of the stored-window target from the
        effect of moving the query off the attractor.
        }
        \label{fig:heldoutfloor}
    \end{subfigure}
    
    \begin{subfigure}[b]{0.8\textwidth}
        \centering
        \includegraphics[width=0.5\linewidth]{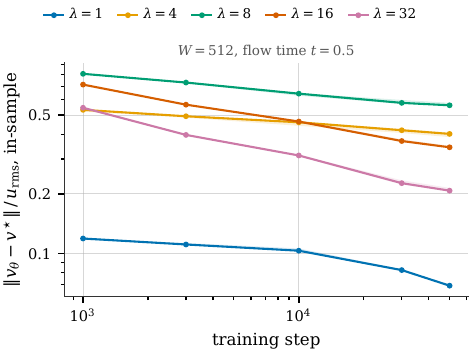}
        \caption{
        \textbf{Training-time convergence toward the closed-form minimiser.}
        Median in-sample fidelity
        $r=\|v_\theta-v^\star\|/u_{\rm rms}$ for the width-$512$ network at
        $t=0.5$, shown across training checkpoints and concentration levels
        $\lambda=\widehat\kappa/\log M$.
        Fidelity continues to improve through the final $50{,}000$-step checkpoint,
        so the reported finite-capacity results should not be interpreted as
        fully converged network limits.
        }
        \label{fig:trainingconvergence}
    \end{subfigure}
    
    \caption{Additional finite-capacity results. Panel separates off-attractor query effects from approximation of the stored window targets. Panel (b) shows fidelity of the trained models continuing to improve at 50000 steps}
    \label{fig:main}
\end{figure}
\FloatBarrier

\begin{figure}[hbtp]
    \centering
    \includegraphics[width=0.9\linewidth]{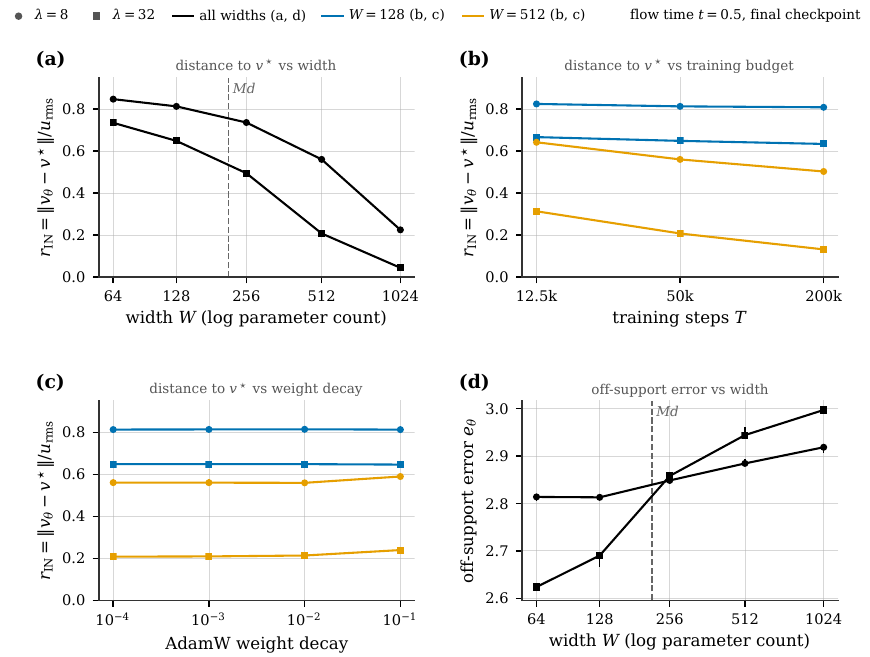}
    \caption{\textbf{Width dependence persists across training budgets and weight decay} (a)$r_{\rm IN}$ against parameter count, widths 64 to 1024, with the baseline of 50k steps and no weight decay. Dashed line: $Md$, the size of the memory bank.(b)$r_{\rm IN}$ against the number of training steps.(c)$r_{\rm IN}$ against AdamW weight decay, at 50k steps.(d)Off-support error $e_\theta$ against the true velocity, against parameter count.}
    \label{fig:ablation}
\end{figure}
\FloatBarrier

\paragraph{Width ablation study design}
Simulated Lorenz-96 system, d=64 with memory bank of M=2000 transitions (evaluation seed 0). The flow is linear, with $\sigma$ held constant in $t$ and set by $\sigma^2=d\hat V/(\lambda \log M) $. An MLP model with three hidden layers of width W with SiLU activations and a linear output of dimension d. Pytorch default initialisation seeded by 1000000+seed and independent of T. Adam optimiser with $\beta=(0.9,0.999)$ or AdamW when weight decay is above 0. Batch size 512, gradient clipping at 1.0. The learning rate follows a linear warmup over 500 steps to 1e-3, the cosine decay to 1e-5 over T steps. Training steps $T\in\{12.5k,50k,200k\}$ for 128 and 512, widths $W\in\{64,128,256,512,1024\}$ and AdamW weight decay $\in\{0,1e-4,1e-3,1e-2,1e-1\}$ for 128 and 512. Every cell is run at $\lambda\in\{8,32\}$ with three seeds. Checkpoints evaluated at $\{1k,3k,10k,30k,50k,100k,200k\}$ upto T plus T itself. 500 fixed queries per set, identical across runs. The median over 500 queries in each run is reported and the [min,max] range over the three seeds.

$r_{\rm IN}$ is the distance between the trained field and $v^\star$ on training-law queries; lower values mean the network is closer to replay. The width dependence persists across the tested configurations: wider networks approach the empirical minimiser more closely, while the width sweep shows higher off-support error for wider models. This supports the robustness of the observed trend, without establishing that optimisation and capacity effects are fully separable.

\subsection{Concentration along generated trajectories} We examine the empirical posterior weights along trajectories generated by the closed-form minimiser $v^\star$ and the trained MLPs of widths $W\in{128,512}$, using the Lorenz-96 configuration of Section 6.4 with $\lambda=\widehat\kappa/\log M=8$.

We consider three initial-query families: training-law samples, held-out states and off-support states. Each field is numerically integrated using Euler integration with 40 steps per unit flow time, and the resulting trajectories are evaluated over 40 autoregressive rollout steps.

At each visited state $x_t$, we compute the posterior responsibilities $\alpha_j(t,x_t)$ of the closed-form empirical estimator, and evaluate $\left(\sum_{j=1}^{M}\alpha_j(t,x_t)^2\right)^{-1}.$

The ESS reported for an MLP-generated trajectory is the ESS of the closed-form posterior evaluated at the states visited by that network. It is not an intrinsic weighting mechanism of the neural network.

We record three quantities: (i) ESS throughout a single flow step, (ii) ESS at the start of successive autoregressive rollout steps, and (iii) distance from each generated state to its dominant stored component, normalised by the kernel-width scale $\sigma\sqrt d$.

\begin{figure}[hbtp]
    \centering
    \includegraphics[width=\linewidth]{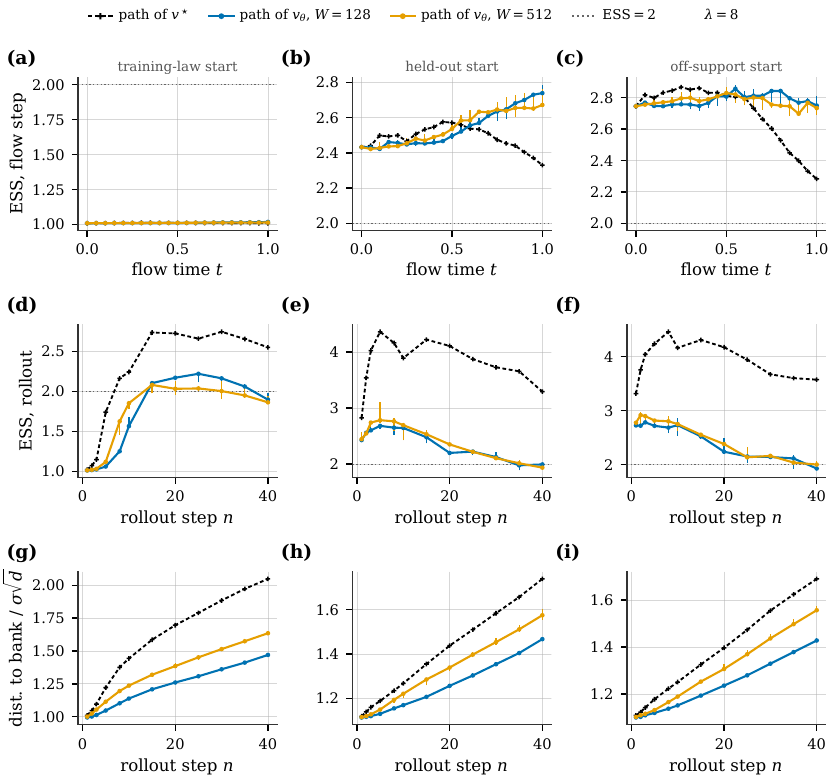}
    \caption{Concentration of the posterior weights along generated paths. Columns represent paths started from training law, heldout windows and off support points. (a--c)ESS along a single flow step.(d--f)ESS at the start of each autoregressive rollout step.(g--i)Distance from current state to dominant stored state, in kernel width units}
    \label{fig:essrollout}
\end{figure}
\FloatBarrier

\cref{fig:essrollout} compares these quantities across the three query families. Training-law starts exhibit nearly one-hot posterior weights, whereas held-out and off-support starts initially exhibit higher ESS. As rollouts progress, all three trajectories move farther from the stored bank. Network-generated trajectories remain closer to the bank than the closed-form trajectories and generally exhibit lower posterior ESS during the rollout.

These observations distinguish posterior concentration from proximity to the training data. A field can remain concentrated on a small number of stored components even when the generated state is far from those components. The experiment is restricted to this Lorenz-96 configuration and does not establish a universal off-support concentration law.

\section{Appendix: Related Works and Future Directions}
\label{app:relworks}

\paragraph{Sequential flow matching} Flow matching and stochastic interpolants have been extended to temporal and forecasting settings through conditional trajectory models, probabilistic forecasting constructions and multi-marginal paths \citep{zhang2024,chen2024si,tamir2024conditional,lim2025paths,lee2025,islam2025,rohbeck2025,kviman2026}. Coupling design has also been studied as a means of simplifying or straightening learned flows \citep{pooladian2023,tong2024}. Our question is different: we hold the empirical regression objective itself under examination and ask when its unrestricted minimiser concentrates onto stored sequential targets

\paragraph{High dimensional nonparametric estimation} The closed-form sequential estimator is a Nadaraya-Watson regression estimator, placing replay within the classical curse of dimensionality for local regression. High-dimensional distance concentration makes nearest neighbour geometry degenerate \citep{beyer1999}, while nonparametric methods can instead adapt to intrinsic dimension when the data possess suitable low-dimensional structure \citep{bickel2007local,kpotufe2011k}. Our covariance result gives an analogous replacement of ambient dimension by a spectral effective dimension for the replay-concentration mechanism.

\paragraph{Memorisation and finite-capacity generalisation} Empirical and theoretical work on diffusion models documents both training-data replication \citep{carlini2023extracting,somepalli2023diffusion} and collapse or memorisation phenomena in the exact generative dynamics
\citep{biroli2024,achilli2024,achilli2025memorizationgeneralizationgenerativediffusion}.
Conversely, learned score networks can exploit geometric or low-dimensional structure
and depart substantially from the corresponding empirical estimator
\citep{kadkhodaie2024,chen2023lowdim,boffi2025}. Recent work also emphasises that finite neural networks need not realise the exact finite-data regression solution \citep{latourelle2026generalisation,bonnaire2025diffusionmodelsdontmemorize}. Our finite-capacity experiment is consistent with this distinction: the replay theory characterises the empirical target, while trained networks may approximate that target only partially.

\paragraph{Future Work} The work done here naturally leads to a few open questions. First, a systematic finite-capacity forecasting study could compare the modelling choices analysed here including window order, interpolation, bandwidth and covariance geometry. 

Another direction is in the theoretical aspects, as discussed in the Discussion and Limitations section where extending the theory for coverage results which do not depend on constant-in-flow-time covariance as well as off-support query log weight gap generalisations for inference could be extremely useful. 

Finally, the finite-capacity results motivate studying how replay emerges during
optimisation. Inspired by \citet{latourelle2026generalisation}, one possible approach
is to analyse network training through a continuous optimisation-time or gradient-flow
limit, with the empirical infinite-capacity solution representing the limiting
regression target. Such an analysis could help characterise when, during training and
as a function of model capacity, a learned network begins to inherit the replay
behaviour of the empirical minimiser.